\documentclass[10pt,twocolumn,letterpaper]{article}
\usepackage[pagenumbers]{cvpr}
\usepackage[dvipsnames]{xcolor}

\usepackage{xcolor}
\usepackage{epigraph} 
\usepackage{bm}

\usepackage{comment}
\usepackage{amsmath}
\definecolor{gray_bg}{HTML}{BFBFBF}
\usepackage{booktabs}
\usepackage{colortbl}
\usepackage{subfig}

\definecolor{Blue1}{RGB}{214, 235, 245}
\definecolor{Blue2}{RGB}{234, 245, 250}
\definecolor{Blue3}{RGB}{247, 251, 253}
\definecolor{Blue25}{RGB}{243, 249, 252}

\definecolor{DarkBlue}{RGB}{0, 0, 100}
\definecolor{BlueT}{RGB}{0, 0, 180}
\definecolor{Grey}{RGB}{180, 180, 180}
\definecolor{Green}{RGB}{0, 255, 0}
\definecolor{Red}{RGB}{255, 0, 0}

\makeatletter
\def\and{
  \end{tabular}%
  \hskip 1em \@plus.17fil%
  \begin{tabular}[t]{c}}
\makeatother

\makeatletter
\def\And{
  \end{tabular}%
  \hskip 1em \@plus.17fil%
  \vskip 0.7em \@plus.17fil%
  \begin{tabular}[t]{c}}
\makeatother

\definecolor{myred}{RGB}{158, 3, 81}
\definecolor{myblue}{RGB}{90, 2, 161}
\definecolor{cvprblue}{RGB}{0.21,0.49,0.74}
\usepackage[pagebackref,breaklinks,colorlinks,citecolor=myblue,linkcolor=myred]{hyperref}

\title{SODA: Bottleneck Diffusion Models for Representation Learning}

\newcommand\blfootnote[1]{
  \begingroup
  \renewcommand\thefootnote{}\footnote{#1}
  \addtocounter{footnote}{-1}
  \endgroup
}

\author{
  Drew A. Hudson$^\star$ \\
  \and
  Daniel Zoran \\
  \and
  Mateusz Malinowski \\  
  \and
  Andrew K. Lampinen \\
  \And
  Andrew Jaegle \\
  \and
  James L. McClelland \\
  \and
  Loic Matthey \\  
  \and
  Felix Hill \\  
  \and
  Alexander Lerchner
}

\begin{document}

\twocolumn[{
\renewcommand\twocolumn[1][]{#1}
\vspace*{-40pt}
\maketitle
\vspace*{-9.2mm}

\begin{center}
\vspace*{-2pt}
\normalsize
    \centering
    \textbf{Google DeepMind}
\end{center}
\begin{center}
    \centering
\vspace*{-1pt}
\centering
\includegraphics[width=1.0\textwidth]{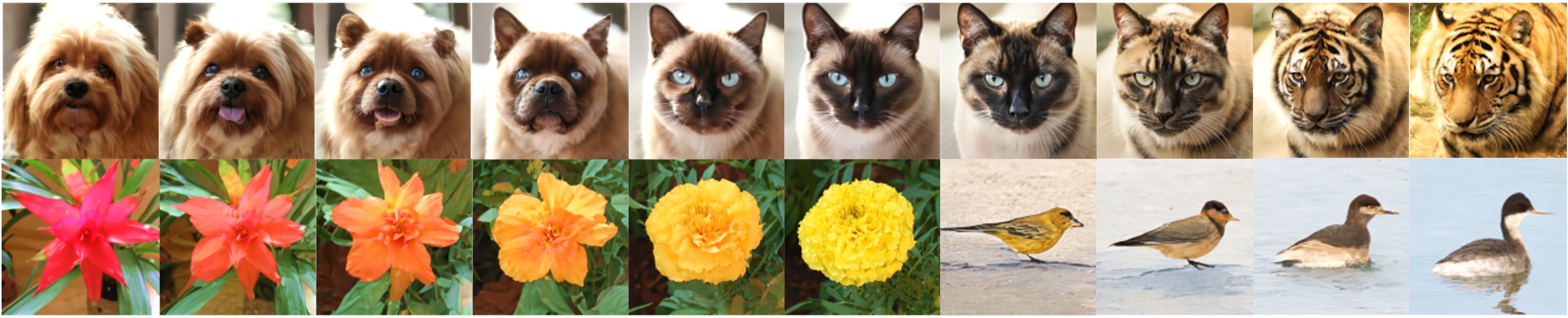}
\vspace*{-17pt}
\captionof{figure}{\textbf{Latent Interpolations.} SODA learns to encode images into compact latent representations. By traversing its latent space, we can interpolate between images, morphing from one image category to another and smoothly  transitioning between semantic attributes.}
\label{fig_interpolation}
\end{center}
}]

\blfootnote{$^\star$Main contributor. The paper presents the findings of Drew's internship project at Google DeepMind, hosted by Felix Hill and Alexander Lerchner. Contact: dorarad@google.com; lerchner@google.com.}

\begin{abstract}
\vspace{-2pt}
We introduce SODA, a self-supervised diffusion model, designed for representation learning. The model incorporates an image encoder, which distills a source view into a compact representation, that, in turn, guides the generation of related novel views. We show that by imposing a tight bottleneck between the encoder and a denoising decoder, and leveraging novel view synthesis as a self-supervised objective, we can turn diffusion models into strong representation learners, capable of capturing visual semantics in an unsupervised manner. To the best of our knowledge, SODA is the first diffusion model to succeed at ImageNet linear-probe classification, and, at the same time, it accomplishes reconstruction, editing and synthesis tasks across a wide range of datasets. Further investigation reveals the disentangled nature of its emergent latent space, that serves as an effective interface to control and manipulate the produced images. All in all, we aim to shed light on the exciting and promising potential of diffusion models, not only for image generation, but also for learning rich and robust representations. See our website at \href{https://soda-diffusion.github.io/}{soda-diffusion.github.io}.
\end{abstract}

\vspace*{-8pt}
\section{Introduction}

\vspace{-6pt}
\epigraph{What I cannot create, I do not understand.}{\textit{-- Richard P. Feynman}}
\vspace{-10pt}

Synthesis, the ability to create, is considered among the highest manifestations of learning \citep{bloom, justin}. As opposed to passive analysis of a text or an image, conceiving them out of thin air involves profound understanding of the underlying factors and intricate generative processes that give rise to the final product \citep{analysis-by-synthesis}. Indeed, learning to write in a new language is often more challenging than reading it. Figuring out the solution to a math problem is fundamentally harder than verifying it \citep{pnp}. And just as the chef learns more about the culinary arts than the diner to prepare a tasty meal, and the novelist knows more about narrative structures than the reader to tell a good story, the artist better grasps perspective and composition to craft a breathtaking masterpiece.

Analogously, in AI, the recent years have witnessed remarkable progress at the generative domain, with large-scale diffusion modeling proving to be a powerful and flexible technique that can create vivid imagery of astonishing realism and incredible detail. And yet, while the vast majority of research harnesses these models for the straightforward goal of synthesis or editing alone \citep{ddpm, diffusion_beat, glide, dalle2, imagen, stable, edit1, edit2, edit3, edit4, edit5, edit6, edit7}, only little attention has been given to their representational capacity \citep{rep1, ddae, rep3}, leaving this promising direction rather unexplored. Surely, models that can weave from scratch such rich depictions of high fidelity, likely learn much along the way about the underlying properties, processes, and components that make up the resulting pictures. How then can we leverage this untapped potential of diffusion models for the purpose of representation learning, and extract the knowledge they acquire for the benefit of downstream tasks?

\begin{figure*}[t]
\vspace*{-10pt}
\centering
  \includegraphics[width=1.0\textwidth]{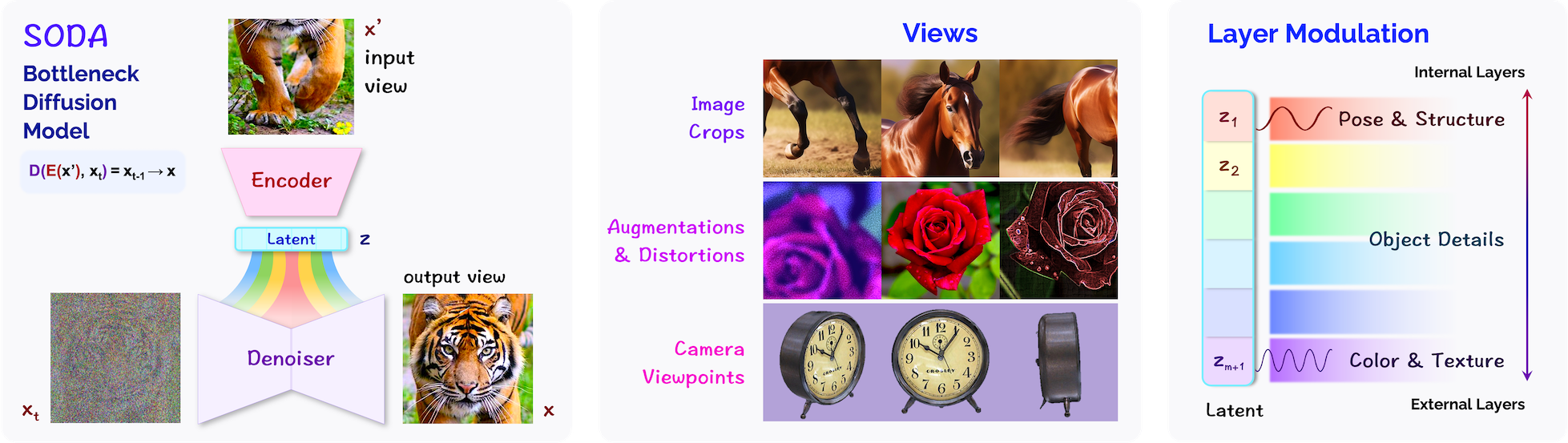}
\vspace*{-15pt}
\caption{\textbf{(Left) Model Overview.} SODA consists of two modules: an \textit{\color{BlueT}  encoder} and a \textit{\color{BlueT}  denoising decoder}. The encoder transforms a source view into a \textit{\color{BlueT} compact latent representation}, which guides the denoising of a target view, by having a dedicated latent subvector modulating each of the decoder's layers \textbf{(Right, Layer Modulation)}. The model is useful both for representation learning, by applying the latents to downstream perception tasks, as well as for synthesis, by iteratively generating novel views. \textbf{(Center) Views.} We consider views as any set of images that relate visually or semantically: They can be crops, augmentations, or images captured from different camera perspectives.}
\vspace*{-7pt}
\label{fig_overview}
\end{figure*}

Motivated to achieve this aim, we present SODA, a self-supervised diffusion model, designed for both perception and synthesis. It couples an image encoder with the classic diffusion decoder \citep{ddpm}, both trained in tandem for novel view generation \citep{nerf} -- a task we choose to employ here, not only for its own sake, but as a self-supervised objective. The encoder transforms an input view into a concise latent representation, which then guides the denoising of an output view, by modulating the decoder's activations. 

This setup introduces a desirable information bottleneck between the encoder and the decoder \citep{bottle}, that in contrast to the typical diffusion framework, equips our model with an explicit and interpretable visual latent space. As our experiments confirm, its advantages are twofold: it both encourages the emergence of disentangled and informative \textbf{representations} that capture image key properties and semantics, which thus can be applied to downstream tasks, and further provides effective means to control and manipulate the produced outputs, for the gain of image editing and \textbf{synthesis}. We further devise and integrate multiple new ideas into the network architecture and training procedure: layer modulation, modified classifier-free guidance, and an inverted noise schedule, so to maximize its representation skills.

We demonstrate our model's strengths and versatility by evaluating it along a series of classification, reconstruction and synthesis tasks, spanning an extensive collection of datasets that covers both the simulated and real-world kinds. SODA possesses strong representation skills, attaining high performance in linear-probing experiments over the ImageNet dataset among others. Moreover, it excels at the task of few-shot novel view generation, and can flexibly synthesize images either conditionally or unconditionally, as indicated by metrics of fidelity, consistency and diversity. Finally, we inspect the model's emergent latent space and discover its disentangled nature, which offers controllability over the semantic traits of the images it produces, as validated both qualitatively and quantitatively. 

Overall, SODA integrates together three research ideas that we seek to establish and promote: First, diffusion models are not only adept at image generation, but are also capable of learning strong representations. Second, novel view synthesis can serve as a powerful self-supervised objective for model pre-training. And third, the compactness of the latent space, which could be reached by constricting the bottleneck between the encoder and the denoiser, plays a pivotal role in enhancing the latent representations' quality, informativeness and interpretability.

\section{Related Work}
\label{related}

\textbf{Diffusion.} The advent of diffusion models has lately marked a breakthrough in the field of visual synthesis. Originally inspired by theories of thermodynamics \citep{thermo}, it approaches generative modeling by following a reversible and iterative denoising process, the forward direction of which slowly erodes the structure within the data distribution, while the backward direction is gradually restoring it. 
Since its early inception back in 2015 \citep{diffusion}, tremendous strides have been made in the quality and diversity of the created outputs, thanks to innovations of the framework's training and sampling techniques \citep{ddim, diffusion_nvidia, classifier, cascading}. Consequently, diffusion models have been widely adopted for numerous tasks and modalities \citep{superres, palette, lion, slotdiff, dorsal}, synthesizing images, videos, audio and text \cite{video1, video2, audio, text1, text2}, and even advancing planning \citep{planning} and drug discovery \citep{drug1}, effectively becoming one of the leading paradigms for generative modeling nowadays.
 
But while most literature highlights its generative feats, only a handful of works have studied diffusion modeling's representational capacity, mainly repurposing pre-trained text-to-image models for classification \citep{rep4}, segmentation \citep{rep2}, or multimodal reasoning \citep{rep5}. The reliance on such models makes it unclear whether the downstream capabilities arise from the diffusion approach itself, or are actually attributable to the exceptionally large scales, long training and voluminous captioned data, which, essentially, provides rich and textual semantic supervision. To address this shortcoming, we focus here instead on the fully-unsupervised regime, and train our model from scratch on standardized benchmarks, seeking to asses the value and potential of diffusion-based representations derived from images alone. 

\textbf{Visual Encoding}. Closer to our work is DRL \citep{rep3}, that extends early research on denoising auto-encoders \citep{dae,dae1,dae2,dae3}, and conditions a denoiser on an encoded clean version of its own target. It is mainly explored from a theoretical perspective, along with preliminary results on MNIST and CIFAR-10. DiffAE \citep{diffae} follows up, integrating style modulation into the encoder \citep{stylegan, film, groupnorm}, while InfoDiffusion \citep{infodiff} regularizes it with mutual-information loss. Our approach builds upon this line of research, but instead of auto encoding the same image, we generate novel views. Notably, we discover that this, in turn, remarkably enhances the model's representation skills, as evidenced by substantial gains in downstream performance. We further couple this idea with multiple technical innovations, pertaining both architecture and optimization, geared to realize the representational capabilities of diffusion models to their fullest. And in contrast to prior works, we provide an extensive empirical study of diffusion-based representation learning, encompassing a broad suite of datasets over multiple different tasks.

\textbf{Hybrid Models}. A couple of partially related works are unCLIP \citep{dalle2} and Latent Diffusion Models \citep{stable}, both of which utilize a frozen pre-trained encoder (CLIP and VQGAN respectively) to cast images onto a compressed latent space over which a diffusion model can operate. Consequently, we note that the latent representations used in both these approaches are in fact not derived by diffusion itself, but rather through either contrastive or adversarial pre-training. As such, they differ fundamentally from our study, which aims to explore the effectiveness of diffusion-based pre-training as a means for representation learning.

\textbf{Downstream Tasks}. For each of the tasks we explore -- classification, disentanglement, reconstruction, and novel view synthesis -- we compare SODA to the leading prior works. These include models such as SimCLR, DINO, and MAE for linear-probe classification \citep{mae, beit, igpt, dino, swav, simclr}, NeRF-based approaches for novel view generation \citep{nerfvae, pixelnerf}, and classic variational models for the task of disentanglement \citep{disen_lib, beta_vae, factor_vae}. Whereas these techniques are designed for particular objectives or depend on domain-specific assumptions, SODA exhibits a greater degree of versatility, as it tackles representational and generative tasks alike.  

\section{Approach}
SODA is a self-supervised diffusion model that learns a bidirectional mapping between images and latents. It consists of an image encoder $\mathcal{E}(\bm{x'})=\bm{z}$ that casts an {\color{BlueT} \textit{input view}} $\bm{x'}$ into a low-dimensional latent $\bm{z}$, which is then used to guide the synthesis of a novel {\color{BlueT}\textit{output view}} $\bm{x}$, that relates to the input $\bm{x'}$ (\cref{fig_overview}). Concretely, $\bm{x}$ is produced through a diffusion process that is conditioned on the encoding $\bm{z}$ via \textit{feature modulation} \citep{stylegan}. This design equips SODA with an explicit and compact latent space, which not only offers ample control over the generative process, but can also be leveraged for downstream perception tasks (\cref{exps})\footnote{Our model is named after the soda drink. Indeed, the fizzing in soda bottles is an everyday example of the diffusion phenomena.}. 

We first present an overview of the model (\textbf{\cref{overview}, \cref{fig_overview}}), followed by an in-depth discussion of each of its core components: the encoder's architectural design (\textbf{\cref{encoder}}), the mechanisms involved in the synthesis of novel views (\textbf{\cref{views}}), and the optimization techniques we develop to cultivate strong and meaningful representations (\textbf{\cref{training}}).

\begin{figure}[t]
\scriptsize
\centering
\includegraphics[width=1.0\linewidth]{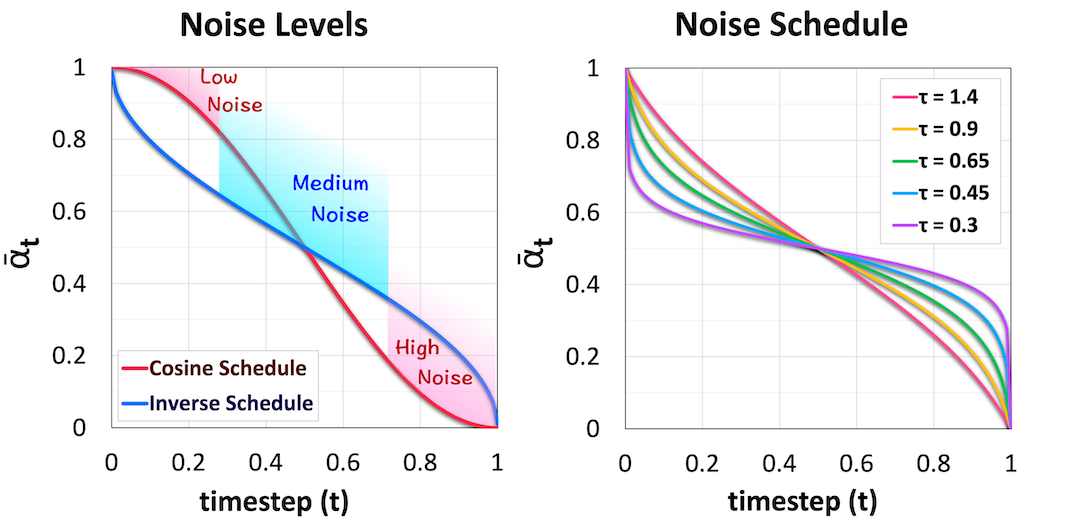}
\vspace*{-10pt}
\caption{\textbf{Inverted Noise  Schedule} used for SODA's training. It prioritizes medium noise levels, which aid representation learning.}
\label{fig_noise}
\vspace*{-10pt}
\end{figure}

\begin{figure*}[t]
\vspace*{-13pt}
\centering

\subfloat{\includegraphics[width=0.247\linewidth]{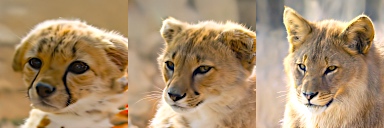}}
\hfill
\subfloat{\includegraphics[width=0.247\linewidth]{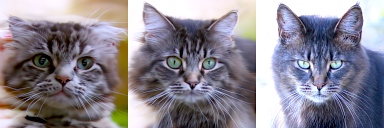}}
\hfill
\subfloat{\includegraphics[width=0.247\linewidth]{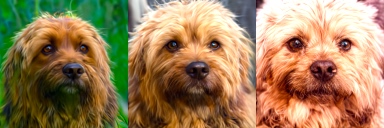}}
\hfill
\subfloat{\includegraphics[width=0.247\linewidth]{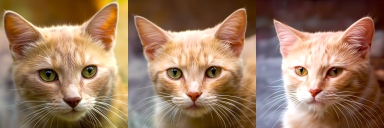}}
\vspace*{-9.1pt}
\subfloat{\includegraphics[width=0.247\linewidth]{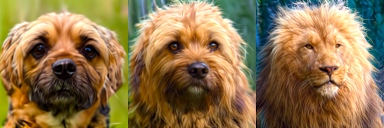}}
\hfill
\subfloat{\includegraphics[width=0.247\linewidth]{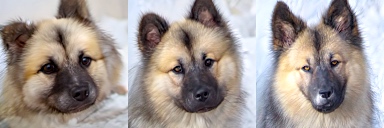}}
\hfill
\subfloat{\includegraphics[width=0.247\linewidth]{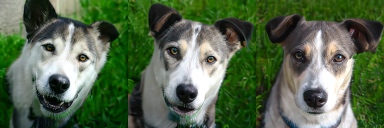}}
\hfill
\subfloat{\includegraphics[width=0.247\linewidth]{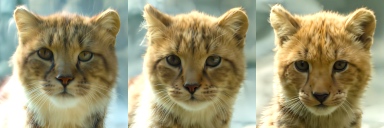}}
\vspace*{-9.1pt}
\subfloat{\includegraphics[width=0.247\linewidth]{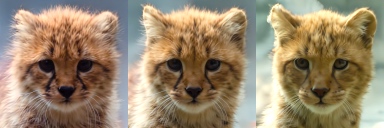}}
\hfill
\subfloat{\includegraphics[width=0.247\linewidth]{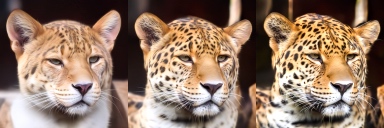}}
\hfill
\subfloat{\includegraphics[width=0.247\linewidth]{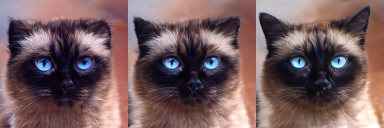}}
\hfill
\subfloat{\includegraphics[width=0.247\linewidth]{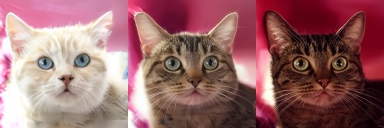}}
\vspace*{-9.1pt}
\subfloat{\includegraphics[width=0.247\linewidth]{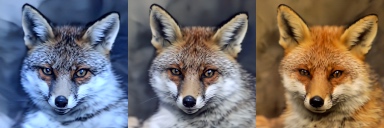}}
\hfill
\subfloat{\includegraphics[width=0.247\linewidth]{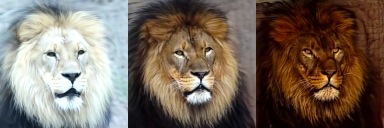}}
\hfill
\subfloat{\includegraphics[width=0.247\linewidth]{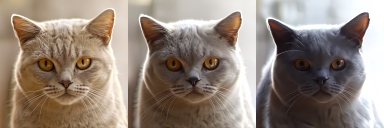}}
\hfill
\subfloat{\includegraphics[width=0.247\linewidth]{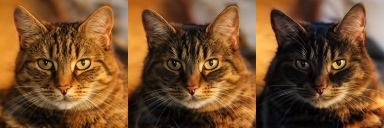}}
\vspace*{-3pt}
  \caption{\textbf{Latent Controllability (Unsupervised).} We explore SODA's latent space and discover directions that correspond to semantic attributes like face size and structure, lighting and viewpoint, maturity, expressions, fur length and color, texture, brightness and tone. SODA is trained fully unsupervised over images only. It successfully disentangles and encodes meaningful aspects into its learned representations.} 
\vspace*{-8pt}
\label{fig_controllability}
\end{figure*}

\subsection{Model Overview} 
\label{overview}
As a denoising diffusion model \citep{ddpm, ddim}, SODA is formally defined by a pair of forward and backward Markov chains, iteratively transforming a sample $\bm{x}_T$ from the normal distribution into the target one ($\bm{x}_0$) and vice versa. Each forward step $t$ erodes $\bm{x}_t$ by adding low Gaussian noise $\bm{\epsilon}_t$ according to a fixed variance schedule $\alpha_t$. Meanwhile, the respective backward step performs image denoising, and aims to estimate $\bm{\epsilon}_t$ in order to recover $\bm{x}_{t-1}$ from its successor $\bm{x}_t$. It is carried out by a decoder $\mathcal{D}$, implemented as a convolutional UNet \citep{unet} (with $2m+1$ activation layers $h_i$).

To tackle the denoising challenge, we assist the decoder by conditioning it on a latent vector $\bm{z}$, which guides its operation by modulating the activations $\bm{h}$ at each of its layers through adaptive group normalization that controls their scale and bias: $\bm{z}_s\text{GroupNorm}(\bm{h})+\bm{z}_b$ where $(\bm{z}_s, \bm{z}_b)$ are linear projections of $\bm{z}$, applied evenly across the activation grid. See Appendix B for the architectural details of the decoder, as well as closed-form equations of the diffusion forward and backward steps. The latent $\bm{z}$ is created by the newly introduced image encoder $\mathcal{E}$, discussed next.

\subsection{The Encoder}
\label{encoder}

The latent representation $\bm{z}$ is at the model's core, serving as the communication channel between the encoder $\mathcal{E}$ and the denoising decoder $\mathcal{D}$, while guiding the latter through the diffusion process. It is derived by a ResNet encoder $\mathcal{E}(\bm{x'})=\bm{z}$, representing a clean source view $\bm{x'}$ that semantically or visually relates to the target view $\bm{x}$
(\cref{views}).

The driving motivation behind this idea stems from the denoising task's inherent \textit{under-determination}: Seeking to fulfill it to the best of its ability, the decoder will leverage any pertinent knowledge or useful clue that could inform it of the missing content to fill in. This in turn incentivizes the encoder to distill into $\bm{z}$ the most striking and prominent commonalities between the source and target views. Since we further constrict the encoding $\bm{z}$ into a \textbf{low-dimensional space}, and use it to guide the decoder via \textbf{global feature modulation}, it yields a fruitful combination that encourages $\bm{z}$ to specifically capture the image's \textit{\color{BlueT} {high-level semantics}}, while delegating the reconstruction of \textit{\color{BlueT} {localized and high-frequency details}} to the denoiser itself.

From that perspective, learning a latent $\bm{z}$ that supports image \textbf{denoising}, rather than pure reconstruction from scratch, as in most auto-encoders \citep{auto-encoders, vae, beta-vae}, liberates our encoder from the need to compress \textit{all} information about the image into the representation, and let it instead focus on the image's most distinctive and descriptive qualities.
\vspace*{-6pt}
\subsubsection{Layer Modulation \& Masking}
\label{masking}

To enhance the model's latent space disentanglement, we introduce two intertwined mechanisms of modulation and masking. In \textbf{layer modulation}, we partition the latent vector $\bm{z}$ into $m+1$ sections -- half the number of layers in the decoder $\mathcal{D}$. We use each $\bm{z}_i$ to modulate the respective pair or layers $(h_i,h_{2m-i})$, thereby promoting specialization among the latent sub-vectors. Just like light rays refracted through a prism, they are encouraged to capture visual traits at different levels of granularity, from coarser to finer, so to guide the decoder's operation through the layers.

To improve the localization and reduce the correlations among the $m$ sub-vectors, we present \textbf{layer masking} -- a layer-wise generalization of classifier-free guidance \citep{classifier}. During training, we zero out a random susbset of $\bm{z}_{1:m}$, effectively performing layer-wise guidance dropout, that mitigates the decoder's reliance on sub-vector dependencies, allowing them to decouple and specialize independently. At sampling, we extrapolate the model's output in the conditional direction: $\bm{\epsilon}_\theta(\bm{x}_t{\mid}\bm{z})-\bm{\epsilon}_\theta(\bm{x}_t{\mid}\bm{0})$. This endows SODA with finer control over the generative process, and opens the door for image editing and style mixing \citep{stylegan}, as we can selectively condition the decoder on some levels of granularity, like structural or positional aspects, while giving it free rein to unconditionally vary other ones, such as lighting, texture, or color palette (see supplementary figures).

\subsection{Novel View Generation}
\label{views}
We loosely consider {\color{BlueT} \textit{views}} to be any set of images that hold some relation among each other, such as visual or semantic (\cref{fig_overview}): they can be various augmentations or distortions of an original image, as is commonly explored in the contrastive learning literature \citep{simclr}, they can show a 3D object from different poses and perspectives \citep{nerf}, or they can simply share the same semantic category with one another. 

\textbf{Cardinality.} We permit the trivial singular case where all views are identical, which then turns the model into an auto-encoder. Conversely, we can extend the conditioning and create a novel view based on a \textit{set of $k$ input views}, instead of just a single one. We map each input $x^i$ to its latent with a shared encoder $\mathcal{E}$, and aggregate the resulting latents $\bm{z}^{1:k}$ into a single vector $\bm{z}$, either by taking their mean, or by processing them through a shallow transformer. 
 
\textbf{Perspective.} The model can incorporate richer forms of conditional information, such as the camera perspective associated with each view: Specifically, for experiments over 3D datasets like ShapeNet (\cref{nvs}), we concatenate a grid of ray positions and directions $\bm{r}=(\bm{o},\bm{d})$, embedded with sinusoidal positional encoding \citep{transformer}, to the linearly-mapped RGB channels of the  source and denoised views, $\bm{x'}$ and $\bm{x}_t$. This allows us to conditionally generate novel views that match the requested pose and orientation. See supplementary for illustrations and implementation details.

\textbf{Guidance.} We extend classifier-free guidance for novel view synthesis, and instead of masking just the latent $\bm{z}$, we randomly and independently mask either the latent or the pose information $\bm{r}$. As our empirical findings suggest, this idea not only hones the model's generative skills, but further enables conditioning on partial information, allowing SODA to either unconditionally conceive \textit{novel objects} at a requested pose, or alternatively, generate arbitrary novel views of given objects at the absence of source or target positional information, based on an image only (Appendix G). 

\textbf{Cross Attention.} The technique of layer modulation offers the encoding $\bm{z}$ with \textit{global control} over each of the decoder's layers, by evenly setting their scales and biases across the grid. We further study alternative mechanisms, and explore the integration of cross attention, so to support \textit{spatial modulation}. Instead of layer-wise modulation (which we use by default), we partition the latent $\bm{z}$ into $n$ sub-vectors, and  perform cross attention between the sub-vectors set and the decoders' activations, akin to the word-based attention common in text-to-image generation. We find that cross attention aids the model at 3D novel view synthesis, while layer modulation performs better for image editing, reconstruction, and representation learning.

\subsection{Training \& Sampling}
\label{training}
\textbf{Noise Schedule.} We train the model with the standard MSE objective \citep{ddpm}, but introduce a new noise schedule to better fit the representation learning task: Indeed, diffusion models commonly set the variance of the additive noise term $\epsilon$ to follow either a cosine \citep{improved_diff}, sigmoid \citep{diff_sig} or linear \citep{ddpm} decay schedules, prioritizing noise levels that are close to the margins, either of the high or low ends (\cref{fig_noise}). Those schedules have been found useful for image synthesis. 

However, from a representation learning perspective, denoising images with overly high or low noise levels fails to provide effective training signal for the model to learn from: Too little noise does not present the denoiser with a challenging enough task, thereby diminishing the encoder's necessity. Meanwhile, too heavy noise puts excess pressure on the latent $\bm{z}$ to fully capture every pixel-level detail of the image, turning denoising into mere reconstruction. We thus incorporate a new inverted noise schedule, that promotes medium noise levels in lieu of the extremes, which proves highly conducive to representation quality (\cref{cls}). 

\textbf{Additional Settings.} Two modifications we find beneficial for representation learning are: (1)~adding low Gaussian noise to the \textit{encoder's input images}; (2)~optionally setting the encoder to have a higher learning rate than the decoder, so to positively impact their learning dynamics by allowing the encoder to adapt faster as it guides the decoder in the denoising task. While the model is robust to the selection of the learning-rate ratio, tuning it could improve downstream results. Once trained, we use DDPM \citep{ddpm} for sampling.

{
\setlength{\tabcolsep}{5pt}
\begin{table}[t]
\caption{\textbf{Linear-Probe Classification on ImageNet}, evaluating discriminative and generative approaches of comparable model sizes. \textit{Top1 (Crop+Flip)} presents scores with light augmentations only, revealing the greater robustness of generative approaches in general and SODA in particular to data augmentations. ($\ast$) denotes cropping only, and ($\dagger$) denotes no augmentation. ($\star$) indicates that the decoder is not used for the downstream linear probing.}
\vspace*{-2pt}
\label{tab_cls}
\centering
\setlength{\tabcolsep}{4.7pt}
\scriptsize
\begin{tabular}{lccccc}
\rowcolor{Blue1} {\textbf{Method}} & {\textbf{Arch.}}  & {\textbf{\#$\theta$}} & {\textbf{Top1}} & {\textbf{Top5}} & {\textbf{Top1}}  \\ 
\rowcolor{Blue1}  & & & &  & {\textbf{Crop+Flip}} \\ 
\rowcolor{Blue2} \multicolumn{6}{l}{\color{DarkBlue} {\textbf{Discriminative Approaches}}} \\
\rowcolor{Blue3} {Supervised} \citep{resnet} & RN50$\times$2 & 94 & 79.9 & 95.0 & - \\ 
SimCLR \citep{simclr} & RN50$\times$2 & 94 & 74.2 & 92.0 & 46.7 \\
BYOL \citep{byol} & RN50$\times$2 & 94 & 77.4 & {\textbf{93.6}} & 63.8 \\
SwAV \citep{swav} & RN50$\times$2 & 94 & 73.5 & - & 54.2 \\
DINO \citep{dino} & ViT-B/16 & 86 & 74.9 & - & 61.1 \\
SwAV + multi-crop & RN50$\times$2 & 94 & 77.3 & - & 58.7 \\
\rowcolor{Blue3} DINO + multi-crop & ViT-B/16 & 86 & {\textbf{78.2}} & - & \textbf{65.3} \\
\rowcolor{Blue2} \multicolumn{6}{l}{\color{DarkBlue} {\textbf{Generative Approaches}}} \\
Vanilla EncDec & RN50$\times$2 & 118 & 8.5 & 17.6 & 10.2 \\
Vanilla AutoEnc & RN50$\times$2 & 118 & 14.3 & 28.9 & -  \\
iGPT \citep{igpt} & GPT-2 & 76 & 41.9 & - & 41.9$^\ast$ \\
iGPT-L \citep{igpt} & GPT-2 & 1386 & 65.2 & - & 65.2$^\ast$ \\
BEiT \citep{beit} & ViT-B/16 & 86 & 56.7 & - & - \\
BigBiGAN \citep{bigbigan} & RV50$\times$4 & 86 & 61.3 & 81.9 & 61.3$^\dagger$ \\
\rowcolor{Blue3} MAE \citep{mae} & ViT-B/16 & 86 & 68.0 & - & {68.0} \\
\rowcolor{Blue2} \multicolumn{6}{l}{\color{DarkBlue} {Diffusion-based Approaches (Generative)}} \\
Palette \citep{palette} & UNet & 118 & 11.4 & 22.3 & 8.7 \\
Unconditional Diffusion & UNet & 118 & 24.5 & 44.4 & 28.3 \\
\rowcolor{Blue3} SODA w/o bottleneck & RN50$\times$2 & 94+35$^\star$ & 34.2 & 52.9 & 29.7 \\
\rowcolor{Blue3} SODA w/o modulation & RN50$\times$2 & 94+32$^\star$ & 56.7 & 78.5 & 51.1 \\
\rowcolor{Blue3} SODA w/o novel views & RN50$\times$2 & 94+36$^\star$ & 55.1 & 75.5 & 48.2 \\
\rowcolor{Blue3} SODA w/o noise sched. & RN50$\times$2 & 94+36$^\star$ & 62.0 & 82.9 & 56.8 \\
\rowcolor{Blue1} \textbf{\color{BlueT} SODA (\textit{ours})} & RN50$\times$2 & 94+36$^\star$ & \textbf{72.2} & \textbf{90.5} & {\textbf{69.1}} \\
\end{tabular}
\end{table}
}

\begin{table}[t]
\caption{\textbf{Image Reconstruction on ImageNet,} comparing discrete \citep{dalle,vqgan} and continuous \citep{dalle2} approaches. The suffix of discrete methods refers to their codebook cardinality.}
\vspace*{-2pt}
\label{tab_recon}
\centering
\scriptsize
\begin{tabular}{lccccc}
\rowcolor{Blue1} \textbf{Method} & \textbf{Latent Dim} & \textbf{PSNR $\uparrow$} & \textbf{SSIM $\uparrow$} & \textbf{FID $\downarrow$} & \textbf{LPIPS $\downarrow$} \\
{DALL-E2} & {1024} & 9.0 & 0.11 & 16.53 & 0.66 \\
\rowcolor{Blue25} DALL-E,8K & 512${\times}$32${\times}$32 & 22.8 & 0.73 & 32.01 & 1.95 \\
{VQGAN,1K} & 256${\times}$16${\times}$16 & 19.4 & 0.50 & 7.94 & 1.98 \\
\rowcolor{Blue25} {VQGAN,16K} & 256${\times}$16${\times}$16 & 19.9 & 0.51 & 4.98 & 1.83 \\
{VQGAN,8K} & 256${\times}$32${\times}$32 & 22.2 & 0.65 & \textbf{1.49} & 1.17 \\
\rowcolor{Blue1} \textbf{\color{BlueT} SODA (\textit{ours})} & {2048} & \textbf{23.6} & \textbf{0.93} & 2.77 & \textbf{0.19} \\
\end{tabular}
\vspace*{-7pt}
\end{table}

\begin{figure*}[t]
\vspace*{-6pt}
\centering
\scriptsize
  \includegraphics[width=1.0\linewidth]{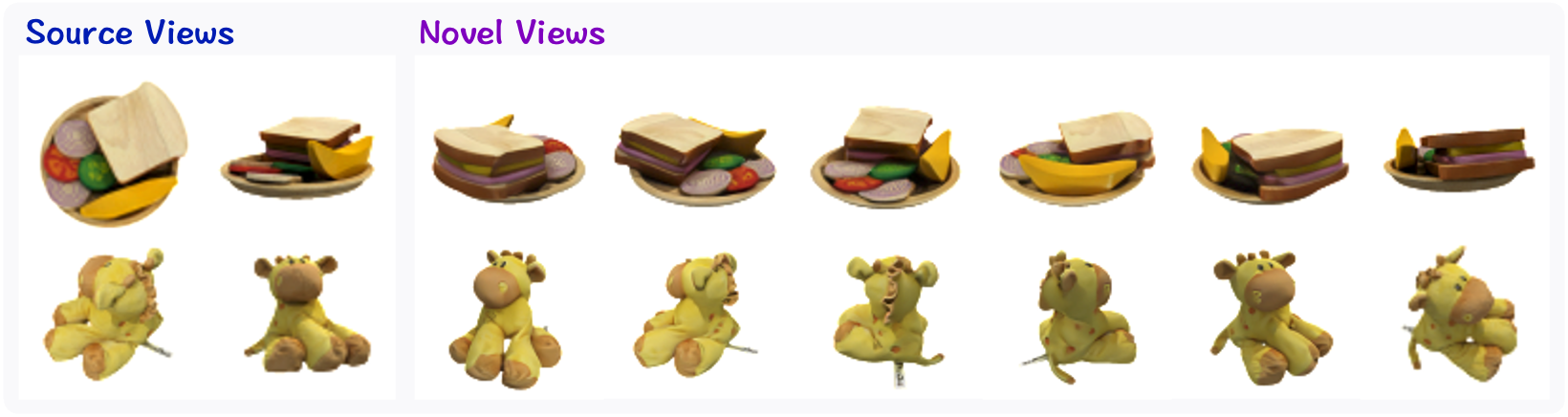}
  \caption{\textbf{Novel View Synthesis}, where given two source views of GSO objects, SODA generates their images from new perspectives.}
\vspace*{-6pt}
\label{fig_nvs_angles}
\end{figure*}

\section{Experiments}
\label{exps}
We evaluate SODA through a suite of quantitative and qualitative experiments, demonstrating its strong representation skills and generative capabilities over 12 different datasets grouped into 4 tasks: We begin with {\color{BlueT} linear-probe classification} (\textbf{\cref{cls}}), showing the model's utility for downstream perception. We proceed to {\color{BlueT} image reconstruction} and few-shot {\color{BlueT} novel view synthesis} (\textbf{\cref{synth}}), illustrating its ability to envision 3D objects from new unseen perspectives. We then explore the model's {\color{BlueT} disentanglement \& controllability} (\textbf{\cref{disen}}), as substantiated by comparative analysis and latent-space interpolations.

In the supplementary and our website website (\textit{\href{https://soda-diffusion.github.io/}{soda-diffusion.github.io}}), we provide additional samples, visualizations and animations, and provide further details about our evaluation procedures, discussing the {\color{BlueT} datasets, metrics, baselines, and implementation details} (\textbf{Appendices C-F}). We conclude with {\color{BlueT} ablation studies} (\textbf{Appendix G}), that empirically validate the contribution of the model's components and design choices. Taken altogether, the evaluation offers solid evidence for the efficacy, robustness and versatility of our approach.

\subsection{Linear-Probe Classification}
\label{cls}
We assess the quality of SODA's learned representations through linear-probe analysis \citep{cls1,cls2,cls3,byol}, over ImageNet and CelebA, which complement one another:
the former calls for fine-grained clustering into 1000 possible \textit{categories}, while the latter involves rich identification of diverse semantic \textit{attributes}. We train our model in a self-supervised fashion: using RandAugment \citep{rand_augment} for ImageNet, and Gaussian data augmentation for CelebA. We then fit a linear classifier over the latent vectors $\bm{z}$ that predicts the category or attributes, and measure the resulting performance. We note that training diffusion models for representation learning is computationally efficient, since iterative sampling is necessary for generative purposes only.

As shown in \cref{tab_cls}, SODA reaches \textbf{72.24\%} accuracy (\textbf{{top1}}) on the ImageNet1K linear-probe classification task, outshining competing generative approaches such as MAE, BEIT and iGPT \citep{mae, beit, igpt}, and significantly reducing the gap with discriminative and constrastive approaches like DINO, SwAV, and SimCLR \citep{dino, swav, simclr}. Meanwhile, for CelebA, our model attains the strongest results (72.7\% F1) compared to competing approaches (Supp Table 6), eclipsing even the language-supervised CLIP embeddings (71.1\% F1).

SODA proves remarkably \textbf{robust to the choice of data augmentation}, as it performs strongly regardless of the selected strategy, seeing only a minor decrease of 3.1\% when switching the heavier RandAugment for a lighter {\textbf{crop+flip}} augmentation. This stands in stark contrast to the high sensitivity of contrastive methods to data augmentations, with e.g. BYOL and SimCLR suffering from major drops of 13.6\% and 27.5\% respectively when light augmentation is applied (crop+flip), and other approaches relying on particular schemes such as MultiCrop \citep{swav} among others. 

\begin{table*}[t]
\vspace*{-3pt}
\caption{\textbf{Novel View Synthesis}, from a single source view, on real and synthetic objects, comparing Geometry-free and -aware approaches. \textit{PSNR}, \textit{SSIM}, and \textit{LPIPS} respectively express pixel-wise, structural and semantic similarity, while \textit{FID} captures sharpness and realism. For these experiments, we integrate cross attention with feature modulation (\cref{views}). ($\star$) denotes short sampling, with 20 steps only.}
\vspace*{-2pt}
\label{tab_nvs}
\setlength{\tabcolsep}{5.5pt}
\centering
\scriptsize
\begin{tabular}{lcccccccccccc}
\rowcolor{Blue1} & {\color{BlueT} \textbf{GSO}} &  &  &  & {\color{BlueT} \textbf{ShapeNet}} &  &  & & {\color{BlueT} \textbf{NMR}}  &  & & \\
\rowcolor{Blue1} \textbf{Method} & \textbf{PSNR $\uparrow$} & \textbf{SSIM $\uparrow$} & \textbf{FID $\downarrow$} & \textbf{LPIPS $\downarrow$} & \textbf{PSNR $\uparrow$} & \textbf{SSIM $\uparrow$} & \textbf{FID $\downarrow$} & \textbf{LPIPS $\downarrow$}  & \textbf{PSNR $\uparrow$} & \textbf{SSIM $\uparrow$} & \textbf{FID $\downarrow$} & \textbf{LPIPS $\downarrow$} \\
\rowcolor{Blue2} \multicolumn{13}{l}{\color{DarkBlue} {Geometry-Aware Approaches}} \\
PixelNeRF \citep{pixelnerf} & 24.93 & 0.919 & 48.72 & 0.086 & 26.58 & 0.940 & 25.34 & 0.073 & 28.19 & 0.932 & 34.54 & 0.082 \\
NeRF-VAE \citep{nerfvae} & 22.20 & 0.882 & 74.83 & 0.113 & 24.60 & 0.915 & 45.79 & 0.101 & 25.56 & 0.893 & 71.67 & 0.134 \\
\rowcolor{Blue2} \multicolumn{13}{l}{\color{DarkBlue} {Geometry-Free Approaches}} \\
Vanilla Auto-Encoder & 24.46 & 0.941 & 75.96 & 0.129 & 24.71 & 0.943 & 42.23 & 0.096 & 24.22 & 0.948 & 51.42 & 0.116 \\
SRT \citep{srt} & 21.97 & 0.877 & 38.64 & 0.110 & 26.31 & 0.934 & 17.96 & 0.073 & 25.69 & 0.898 & 7.90 & 0.090 \\
\rowcolor{Blue2} \multicolumn{13}{l}{\color{DarkBlue} {Diffusion-based Approaches (Geometry-Free)}} \\
Palette \citep{palette} & 13.42 & 0.672 & 8.49 & 0.199 & 14.44 & 0.582 & 6.93 & 0.177 & 14.10 & 0.609 & 6.03 & 0.212 \\
DALL-E2 \citep{dalle2} & 15.68 & 0.793 & 8.32 & 0.147 & 18.75 & 0.823 & 6.54 & 0.101 & 20.32 & 0.899 & 3.61 & 0.087 \\
\rowcolor{Blue2} SODA w/o bottleneck & 20.97 & 0.926 & 4.39 & 0.071 & 24.31 & \textbf{0.949} & 2.83 & 0.051 & 23.31 & 0.948 & \textbf{0.75} & 0.053 \\
\rowcolor{Blue2} SODA w/o modulation & 21.02 & 0.929 & 4.10 & 0.069 & 25.02 & 0.944 & 2.97 & 0.048 & 25.34 & 0.949 & 0.77 & 0.051 \\
\rowcolor{Blue2} SODA (shorter sampling)$^\star$ & 24.38 & 0.930 & 2.35 & 0.065 & 26.71 & 0.946 & 1.31 & 0.046 & 27.13 & 0.936 & 1.10 & 0.063 \\
\rowcolor{Blue1} \textbf{\color{BlueT} SODA (\textit{ours})} & \textbf{24.97} & \textbf{0.945} & \textbf{1.51} & \textbf{0.054} & \textbf{27.42} & 0.947 & \textbf{0.74} & \textbf{0.039} & \textbf{28.71} & \textbf{0.952} & 0.81 & \textbf{0.048} \\
\end{tabular}
\vspace*{-5pt}
\end{table*}

The newly introduced \textbf{image encoder} plays an instrumental role in the model's downstream performance, enabling a 3x boost over features obtained from an \textit{unconditional diffusion} model \citep{ddae}\footnote{The baseline obtains image encodings by pooling the features of the best-performing denoiser layer (a middle one) over a lightly-noised image.}, which for ImageNet scores 24.49\% only. A significant improvement further arises from the use of \textbf{novel view synthesis as a self-supervised representation learning objective}: Indeed, maintaining a distinction between the model's source and target views yields a 17.12\% increase for ImageNet, compared to when they match (i.e. auto-encoding), even though the same data augmentation is applied in both cases (\cref{fig_plot})\footnote{For auto-encoding, we sample a new augmentation at every training step, but use it both as the source and as the target.}.

Other contributors include the \textbf{compact bottleneck} and \textbf{feature modulation}, which respectively raise accuracy by 38.04\% and 15.51\% for ImageNet, and 12.74\% and 11.25\% F1 for CelebA (Supp Table 11). Designing a new inverted \textbf{noise schedule} that favors medium noise levels in lieu of the extremes likewise strengthen the model's representation capacity, eliciting a 10.25\% increase over ImageNet.

\subsection{Visual Synthesis}
\label{synth}
Next, we analyze the model's generative skills, evaluating its ability to faithfully {\color{BlueT} reconstruct} an image $\bm{x}$ from its latent encoding $\bm{z}$ (\textbf{\cref{recon}})\footnote{To evaluate our model at the task of reconstruction, we keep the source the and the target views the same during training $\bm{x'}=\bm{x}_0$.}, and {\color{BlueT} generate novel views} of 3D objects from requested camera perspectives, given one or more conditional source views (\textbf{\cref{nvs}}). We evaluate the targets and predictions' similarity along multiple dimensions: pixel-wise (PSNR \citep{psnr}), structural (SSIM \citep{ssim}), perceptual (FID \citep{fid}) which accounts for sharpness and realism, and semantic (LPIPS \citep{lpips}) (Appendix E.2).

\vspace*{-6pt}
\subsubsection{Image Reconstruction}
\label{recon}
As indicated by \cref{tab_recon}, SODA produces excellent reconstructions, surpassing competing approaches like VQGAN \citep{vqgan}, StyleGAN2 \citep{stylegan2}, DALL-E \citep{dalle} and unCLIP (DALL-E2) \citep{dalle2}, especially in terms of structural and semantic similarity (SSIM \& LPIPS). Visually speaking, our samples are sharper and crispier than DALL-E's, while being more accurate than StyleGAN2 and unCLIP inversions, perhaps due to their lack of a trainable encoder (see supplementary examples). The results are significant given the order-of-magnitude lower dimensionality of the latents $\bm{z}$ from which we restore the images: 2K for SODA versus 65-524K for VQGAN and DALL-E\footnote{The overall latent dimension is $16{\times}16{\times}256$ -- $32{\times}32{\times}512$.}, illustrating here an advantage of continuous representations over discrete codebooks. 

\vspace*{-6pt}
\subsubsection{Novel View Generation}
\label{nvs}
For the task of few-shot novel view synthesis, we focus on the 3D regime and look into 3 datasets that span both synthetic object renderings and real-world scans of household items (Google Scanned Objects \citep{gso}, custom ShapeNet \citep{shapenet}, and NMR \citep{nmr}). We compare our model to geometry-free and -aware approaches such as PixelNeRF \citep{pixelnerf}, Scene Representation Transformer (SRT) \citep{srt}, and Palette \citep{palette}. We condition the models on 1-9 source views, and test them on held-out validation objects that do not appear in training.

SODA consistently beats the competing approaches across the 3 datasets and for  different numbers of source views, as indicated by FID, SSIM and LPIPS (\cref{tab_nvs} and Supplementary Figure 7). It reaches the largest gains along LPIPS and FID, producing significantly sharper images that better match the source views both structurally and semantically. We observe that settings of 1-3 source views benefit the most from our model, where for the single-source case, it improves FID scores by an order-of-magnitude and often almost halves the LPIPS scores. 
For the GSO dataset, as we increase the source views number, we score a little lower on the pixel-wise PSNR than PixelNeRF, perhaps due to the probabilistic nature of our approach. Yet, in terms of computational efficiency, contrary to the slow and heavy rendering of geometry-aware methods, SODA maintains strong performance with as little as 20 sampling steps.

\begin{figure}[t]
\centering
\scriptsize
\vspace*{-12pt}
\subfloat{\includegraphics[width=0.5\linewidth]{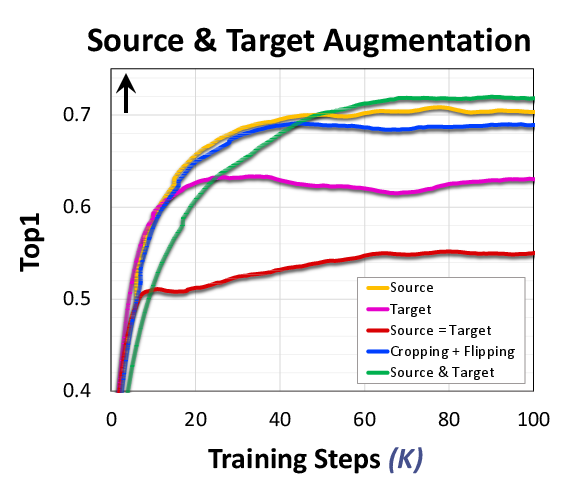}}
\subfloat{\includegraphics[width=0.5\linewidth]{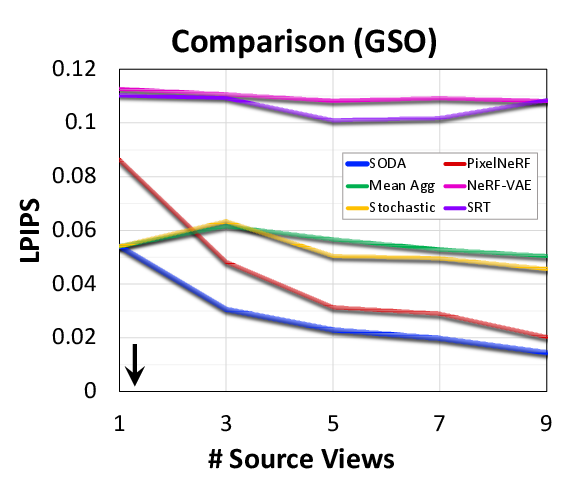}}
\vspace*{-5pt}
\caption{\textbf{(Left):} Contribution of the Novel View Synthesis objective to downstream classification on ImageNet. \textbf{(Right):} Generative quality comparison with baseline approaches and aggregation methods as we increase the number of source views.}
\vspace*{-8pt}
\label{fig_plot}
\end{figure}

\cref{fig_nvs_angles} and the supplementary animations feature objects synthesized from various perspectives, showcasing the viewpoint consistency SODA achieves. For multiple sources, we find that our proposed transformer-based view aggregation (\cref{views}) surpasses the stochastic conditioning technique of 3DiM \citep{nerd} (\cref{fig_plot}). Our approach further outperforms the denoiser-only Palette diffusion model, which fits translational tasks that closely follow the source layout, like colorization or super-resolution, but struggles at structural transformations, corroborating the need for our dedicated image encoder.

\subsection{Disentanglement \& Controllability}
\label{disen}
The concept of disentanglement has been a recurring theme in representation learning research over the years \citep{disen1, disen3, disen4, infogan, dci}. While formal definitions may vary \citep{dis1, dis2, dis3, dis4, disen2}, a common aim lies in the discovery of abstract and meaningful latent representations that linearly align with the natural axes of variation. Disentanglement could enhance the encodings' interpretability, and, in the context of generative modeling, support greater controllability. In the following, we inspect the model's latent space, and analyze it quantitatively and qualitatively along the dimensions of disentanglement, controllability and informativeness.

\vspace*{-6pt}
\subsubsection{Qualitative Evaluation}
\textbf{Latent Interpolations.} We begin by visualizing latent interpolations for our model, linearly traversing the latent space from one vector $\bm{z}_1$ to another $\bm{z}_2$ (\cref{fig_interpolation} and the supplementary figures \& animations). We observe smooth variations over traits of texture and structure. Notable in particular are image categories that seamlessly morph from one to another (e.g. from a tiger to a cat to a dog to a wolf), while for CelebA, we see gradual transformations ranging from broad shifts of pose and orientation to finer transitions of hair, facial features and expressions.

\textbf{Attribute Manipulation.} 
We go beyond interpolations and identify meaningful latent directions that correspond to individual axes of variation. To infer them, we explore two techniques: \textbf{supervised}, by normalizing the linear-probe's weight matrix, and \textbf{unsupervised}, through PCA decomposition \citep{pca_on_stylegan} (details at Appendix E.3). \cref{fig_controllability} and the supplementary figures show perturbations along the discovered directions, which influence various semantic properties: from age and gender in CelebA, to tone, clarity and lighting conditions in LSUN, to object's dimensions, thickness and material in ShapeNet and GSO. Indeed, we see that these manipulations are disentangled: one attribute is altered while the others are mostly kept intact, demonstrating the well-behaved nature of SODA's emergent latent space and the quality and strength of its learned representations. Indeed, we emphasize that both SODA's training and the following PCA-based discovery of interpretable latent directions are fully-unsupervised, derived from images only.

\textbf{Layer Modulation.} We investigate the effect of layer modulation and masking (\cref{masking}), blocking the encoder's guidance from select decoder's layers and inspecting the impact on the produced outputs. As illustrated in the supplementary, it allows for selective modification of input images at different levels of granularity, so to preserve certain factors while unconditionally regenerating other ones. We can thus resample a new color palette while retaining shape and structure, or change the background while preserving the subject's identity. We find that layer masking improves the model's robustness to these forms of partial conditioning. Meanwhile, the role played by the initial Gaussian \textbf{noise map} $\bm{x_T}$ is closely linked to the chosen augmentation scheme: it controls fine stochastic subtleties like fur or freckles when SODA is trained to reconstruct the source image, and could conversely shape the underlying layout when heavier data augmentations are applied.

\vspace*{-6pt}
\subsubsection{Quantitative Evaluation}
\label{disen_quant}
To quantitatively bolster the findings above, we analyze our approach with DCI \citep{dci}, which measures representations along Disentanglement, Completeness and Informativeness by assessing the degree of 1-to-1 correspondence between latent and ground-truth factors of variation (Appendix E.3). We evaluate models over multiple semantically-annotated datasets, ranging from the diagnostic SmallNORB and MPI3D to the realistic CUB and CelebA \citep{smallnorb, mpi3d, shapes3d, cub, celeba}.

\begin{table}[t]
\scriptsize
\caption{\textbf{Disentanglement Analysis (DCI)} for various datasets. SODA achieves improvements of 27.2-58.3\% in Disentanglement, 5.0-23.8\% in Completeness, and comparable Informativeness to variational approaches. We report here the 3 metrics' average. StyleGAN2 achieves 68.71\% (W) and 69.07\% (W+) for CelebA. See Supplementary Table 7 for the full comparison.}
\label{tab_disen}
\centering
\vspace*{-2pt}
\setlength{\tabcolsep}{4.0pt}
\begin{tabular}{lccccc}
\rowcolor{Blue1} & & & & & \\[-6.5pt]
\rowcolor{Blue1} \textbf{Method} & \textbf{\color{BlueT} CelebA} & \textbf{\color{BlueT} CUB} & \textbf{\color{BlueT} MPI3D} & \textbf{\color{BlueT} 3DShapes} & \textbf{\color{BlueT} SmallNORB} \\[0.5pt]
AnnealedVAE \citep{annealed_vae} & 53.09 & 46.94 & 30.95 & 72.11 & 33.14  \\
FactorVAE \citep{factor_vae} & 54.76 & 47.62 & 43.99 & 91.26 & 50.04 \\
\rowcolor{Blue3} DIP-VAE (I) \citep{dipvae} & 58.63 & 46.39 & 55.94 & 94.11 & 47.16 \\
\rowcolor{Blue3} DIP-VAE (II) \citep{dipvae} & 57.02 & 47.22 & 44.40 & 94.07 & 51.20 \\
$\beta$-VAE \citep{beta-vae} & 56.96 & 47.39 & 47.77 & \textbf{94.45} & 50.40 \\
$\beta$-TCVAE \citep{tc_vae} & 56.61 & 48.59 & 50.33 & 88.51 & 50.56 \\
\rowcolor{Blue2} SODA w/o layer mod. & 65.61 & 54.75 & 70.71 & 80.09 & 59.10 \\
\rowcolor{Blue1} \textbf{\color{BlueT} SODA (\textit{ours})} & \textbf{74.67} & \textbf{56.98} & \textbf{73.41} & 94.08 & \textbf{64.78} \\
\end{tabular}
\vspace*{-7pt}
\end{table}

As \cref{tab_disen} and supplementary Tables 6 and 7 show, SODA outshines both variational and adversarial approaches, improving Disentanglement by 27.2-58.3\% and Completeness by 5.0-23.8\% across 4 different datasets, with the sole exception of the synthetic 3DShapes, for which both SODA and most variational methods attain excellent scores. For Informativeness, results are mostly comparable, with SODA taking the lead for some datasets, while StyleGAN or DIP-VAE \citep{dipvae} improving scores for others. Our experiments further validate the contribution of layer modulation and masking, respectively yielding 3.2-13.5\% and 2.5-8.9\% increases in latent-space Disentanglement, and 3.8\% and 1.7\% mean increase in Completeness. Visually, SODA's samples are significantly sharper than the variational ones, and it achieves remarkable boosts in realism (FID) and semantic similarity (LPIPS).

\section{Conclusion}
We introduced SODA, a self-supervised diffusion model, designed for both perception and synthesis. It re-purposes the task of novel view generation as a training objective for representation learning. By conditioning a denoiser on an image encoder, and imposing an information bottleneck between the two, SODA learns strong semantic representations that enable downstream classification, as well as reconstruction, editing and synthesis. While we focused on single-object images, as in LSUN, ShapeNet, or ImageNet, we believe that exploring the applicability of our approach to dynamic compositional scenes is a promising direction for future research. We hope our work will help bridging the gap between novel view synthesis and self-supervised learning, two flourishing topics that are often pursued independently, and bring us one step closer to unlocking the potential of generative models in general and diffusion models in particular to advance the representational frontier.

{
 \small
 \bibliographystyle{ieeenat_fullname}
 \bibliography{main}
}

\clearpage
\newpage
\appendix

\section*{Supplementary Material}
\section{Overview}
In the following, we discuss additional analysis of our approach, and provide further description of the model structure, implementation details, and evaluation procedures. \textbf{\cref{preliminaries}} offers an overview of diffusion models' preliminaries and equations. In \textbf{\cref{supp_impl}}, we then specify the chosen hyperparameters, training techniques, and sampling methods. \textbf{\cref{datasets,eval,baselines}} respectively review the datasets, metrics, and baselines we consider in this study. Finally, in \textbf{\cref{ablt}}, we present ablation and variation studies that assess the contribution of each of our design choices, complementing the principal ones explored in the main paper.

We plan very soon to add to the supplementary and our website (\textit{\href{https://soda-diffusion.github.io/}{soda-diffusion.github.io}}) a variety of animations and visualizations of outputs generated by the model over different datasets, spanning image reconstructions, viewpoint traversals, latent interpolations, unsupervised attribute discovery and manipulation, demonstration of style and content (or structure) separation, qualitative impact of layer masking and variation of the initial noise map for different training data augmentation schemes, and samples conditioned on partial information. 

\section{Model Overview \& Diffusion Preliminaries}
\label{preliminaries}
As a denoising diffusion model \citep{ddpm}, SODA is formally defined by a pair of forward and backward Markov chains that represent a $T$-steps transformation from a normal distribution $\bm{x}_T \sim \mathcal{N}(\bm{0}, \bm{I})$ into the learned data distribution $\bm{x}_0 \sim p_{\theta}(\bm{x})$ and vice versa, where $\bm{x} \,{\in}\, \mathbb{R}^{H\times W \times 3}$. Each forward step $t$ erodes $\bm{x}_t$ by adding a small Gaussian noise according to a fixed variance schedule $\alpha_t$, sampling:
\[\bm{x_{t}} \sim \mathcal{N}(\sqrt{\alpha_t}\bm{x_{t-1}}, \left(1-\alpha_t\right)\bm{I}) \]
Meanwhile, each reverse step $t$ performs image denoising, and aims to estimate $\bm{\epsilon}_t$ in order to recover $p_\theta(\bm{x}_{t-1}\mid\bm{x}_t,\bm{z},\bm{c})$ where the latent representation $\bm{z}\,{\in}\,\mathbb{R}^D$ serves as a guidance source for denoising the image, and is produced by the encoder through $\bm{z}=\mathcal{E}(\bm{x'}, \bm{c'})$, the image $\bm{x'}$ is a related clean input view given to the encoder, and $\bm{c}, \bm{c'}$ denote optional conditions for the encoder and decoder respectively (e.g. source and target camera perspectives of a 3D object). We note that this formulation contrasts with unconditional diffusion models, which rely on $(\bm{x_{t}},t)$ only. The reverse step is realized by a denoising decoder $\mathcal{D}$ that predicts
$\bm{\epsilon}=\bm{\epsilon}_\theta(\bm{x}_t,t,\bm{z},\bm{c})$.
Thanks to the reparametrization trick \citep{ddpm}, we can then sample the following:
\[\bm{x}_{t-1} \sim \mathcal{N}\left(\frac{1}{\sqrt{\alpha_t}}\left(\bm{x}_{t} - \frac{1-\alpha_t}{\sqrt{1-\bar{\alpha}_t}}\bm{\epsilon}_\theta(\bm{\cdot})\right), \sigma_t^2\bm{I}\right) \]
where $\bar{\alpha}_t=\prod\nolimits_{s=1}^t{\alpha_s}$ is the product of the variances up to step $t$, and $\sigma_t^2$ is either a fixed or learned variance term. To train the model, we can readily obtain $\bm{x}_{t}$ with the closed-form computation (where $\bm{\epsilon}\sim\mathcal{N}(\bm{0},\bm{I})$):
\[\bm{x}_{t} = \sqrt{\bar{\alpha_t}}\bm{x}_{t-1} + \left(1-\bar{\alpha_t}\right)\bm{\epsilon}\]
and couple it with the simplified re-weighted MSE training objective (where $\bm{\epsilon}_\theta$ is estimated by the model):
\[\mathcal{L}=\mathbb{E}_{\bm{x}_0,\bm{\epsilon},t}[{\|\bm{\epsilon}-\bm{\epsilon_\theta}]\|}_2^2\]
In terms of the architecture, our model consists of an image encoder $\mathcal{E}$ (ResNet or ViT), and a denoising decoder $\mathcal{D}$ that follows the classic structural design of prior literature \citep{ddpm, diffusion_beat}, featuring a UNet implemented as a stack of residual, convolutional, and either downsampling or upsampling layers (in the encoding and decoding modules of the UNet respectively), that are further linked by symmetric skip connections. The decoder $\mathcal{D}$ notably integrates Adaptive Group Normalization layers \citep{stylegan, film, diffusion_beat} throughout, allowing $\bm{z}$ and $t$ to modulate the decoder's activations of each layer $\bm{h}$, by scaling and shifting them channel-wise:
\[ \text{AdaGN}(\bm{h}, \bm{z}, t) = \bm{z}_s(\bm{t}_s{\text{GroupNorm}(\bm{h})}+\bm{t}_b)+\bm{z}_b \]
where $(\bm{t}_s, \bm{t}_b)$ and $(\bm{z}_s, \bm{z}_b)$ are both obtained by linear projections, the former of a sinusoidal timestep embedding of $t$ \citep{transformer}, and the latter of the latent representation $\bm{z}$ created by the image encoder $\mathcal{E}$.

\begin{table*}[t]
\caption{\textbf{Novel View Synthesis (FID)}, comparing different approaches and aggregation methods as we vary the number of source views. \textit{Stochastic Conditioning} guides each sampling step with a randomly-chosen source view. \textit{Mean Aggregation} conditions on multiple source views by averaging their latents, while \textit{Transformer Aggregation} instead uses a shallow transformer to aggregate the view representations.}
\label{tab_nvs_fid}
\centering
\scriptsize
\begin{tabular}{lcccccccccc}
\rowcolor{Blue1} &  \textbf{\color{DarkBlue}GSO} &  &  &  & & \textbf{\color{DarkBlue}ShapeNet}  &  &  & & \\
\rowcolor{Blue1} \textbf{\# Source Views} & \textbf{1} & \textbf{3} & \textbf{5} & \textbf{7} & \textbf{9} & \textbf{1} & \textbf{3} & \textbf{5} & \textbf{7} & \textbf{9} \\
NeRF-VAE \citep{nerfvae} & 74.835 & 79.984 & 76.965 & 81.334 & 80.926 & 45.791 & 42.165 & 36.592 & 35.441 & 34.134 \\
\rowcolor{Blue2} SRT \citep{srt} & 38.642 & 70.665 & 40.728 & 51.936 & 74.705 & 17.956 & 16.336 & 13.717 & 27.719 & 28.026 \\
PixelNeRF \citep{pixelnerf} & 48.721 & 20.659 & 7.934 & 5.906 & 3.622 & 25.341 & 9.679 & 4.591 & 3.607 & 2.557 \\
\rowcolor{Blue2} SODA with Mean Aggregation & \textbf{1.508} & 2.290 & 2.060 & 2.183 & 1.754 & \textbf{0.736} & 0.696 & 0.667 & 0.679 & 0.742 \\
SODA with Stochastic Conditioning \citep{nerd} & \textbf{1.508} & 2.117 & 1.360 & 1.177 & 1.228 & \textbf{0.736} & 0.706 & 0.686 & 0.679 & 0.697 \\
\rowcolor{Blue1} \textbf{SODA} (Transformer Aggregation) & \textbf{1.508} & \textbf{0.962} & \textbf{0.797} & \textbf{0.711} & \textbf{0.653} & \textbf{0.736} & \textbf{0.491} & \textbf{0.458} & \textbf{0.378} & \textbf{0.319} \\
\end{tabular}
\end{table*}

\begin{figure*}[t]
\centering
\subfloat{\includegraphics[width=0.25\linewidth]{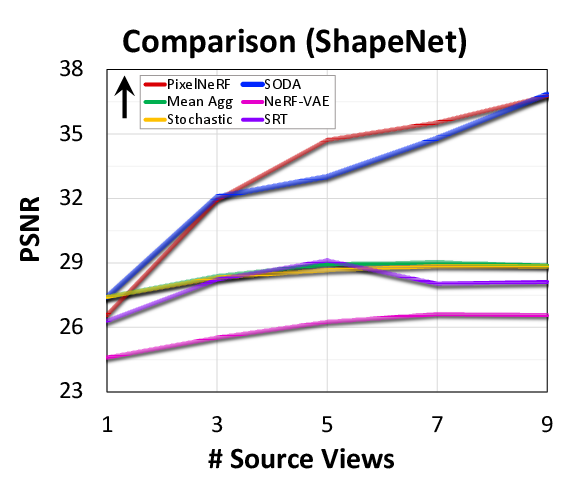}}
\hfill
\subfloat{\includegraphics[width=0.25\linewidth]{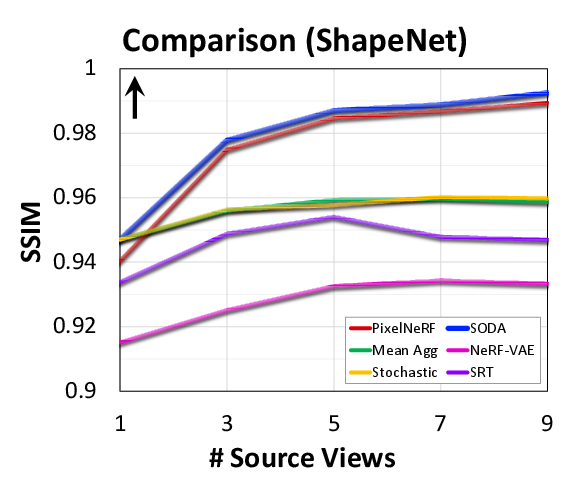}}
\hfill
\subfloat{\includegraphics[width=0.25\linewidth]{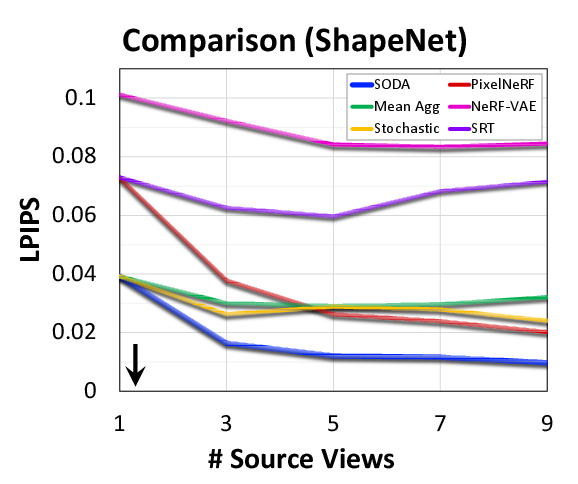}}
\hfill
\subfloat{\includegraphics[width=0.25\linewidth]{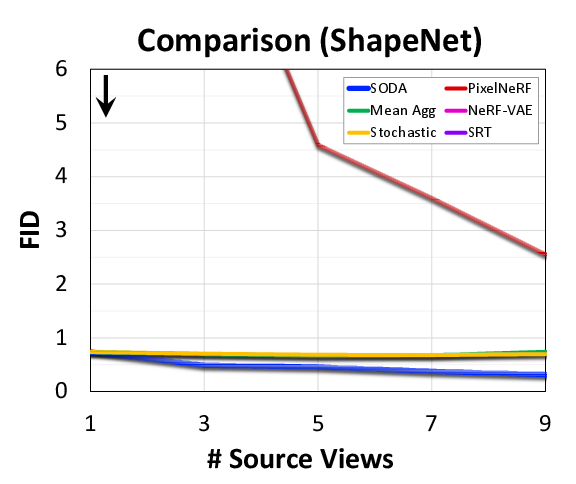}}

\subfloat{\includegraphics[width=0.25\linewidth]{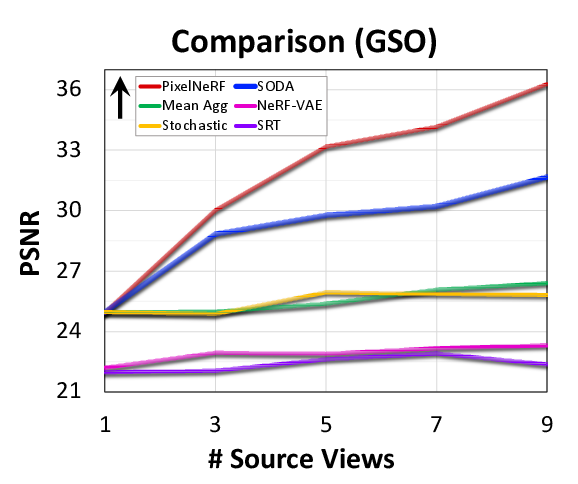}}
\hfill
\subfloat{\includegraphics[width=0.25\linewidth]{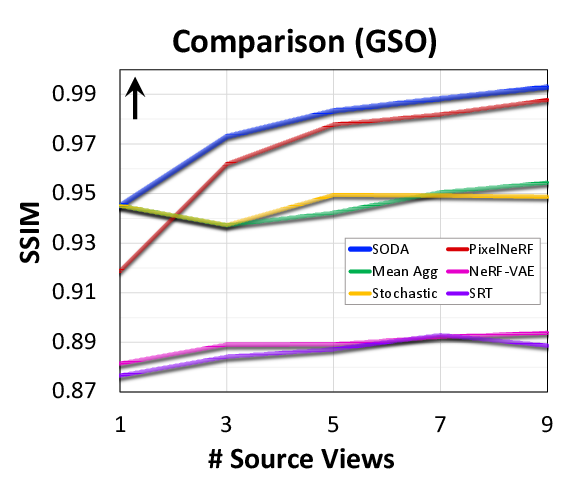}}
\hfill
\subfloat{\includegraphics[width=0.25\linewidth]{plots/gso_lpips.png}}
\hfill
\subfloat{\includegraphics[width=0.25\linewidth]{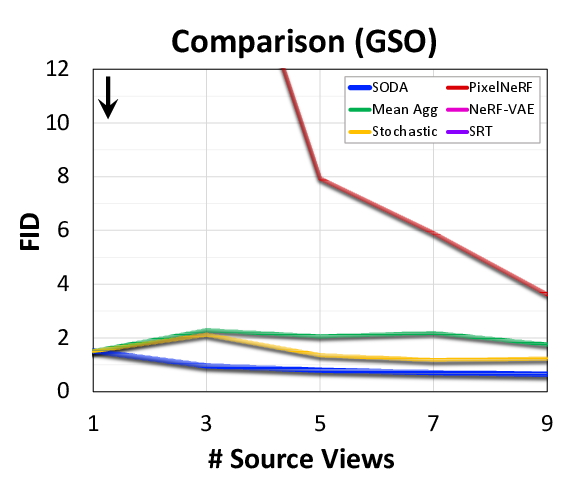}}
\caption{\textbf{Number of source views' impact on models performance}, along PSNR, SSIM, LPIPS and FID. We see that as we increase the number of views, SODA achieves a bit higher SSIM score than PixelNeRF, comparable or lower PSNR, better LPIPS score, and much better FID. Other approaches lag behind PixelNeRF and SODA. In terms of view aggregation, we see that stochastic conditioning performs similarly to averaging the view representations, and that our transformer-based aggregation performs robustly better than these alternatives.}
\label{fig_nvs}
\end{figure*}

\section{Implementation Details} 
\label{supp_impl}
\textbf{Architecture.} See \cref{tab_impl} for our chosen hyperparameters. In terms of the training objective, optimization scheme and empirical configuration, we adopt most of the common settings of recent works \citep{ddpm, diffusion_beat, glide}, and specifically use the Adam optimizer \citep{adam}, gradient accumulation, and exponential moving average for the model's weights; for the ResNet encoder \citep{resnet}: variant v2 ResNet \citep{resnet_v2}, Xavier initialization \citep{xavier}, ReLU non-linearity, dropPath \citep{dropath}, and mean pooling; and for the UNet decoder: truncated normal initialization (JAX default), GeLU non-linearity \citep{gelu}, $\sqrt{2}\,$ rescaling of residual connections, BigGAN re-sampling order \cite{biggan}, and self attention in the decoder's low-resolution layers (8-32).

\textbf{Training.} For each dataset, we train the model until convergence, as measured by lack of improvement over a set number of training steps along a validation metric of choice (either downstream accuracy or SSIM). For sampling, we use discrete-time DDPM \citep{ddpm}, classifier-free guidance \citep{classifier} and 1000 diffusion timestemps, practically strided into 75-250 steps \citep{improved_diff}. We implement SODA in JAX \citep{jax}, and run our experiments either on NVIDIA Tesla V100s or TPUs (v2). 

\textbf{Positional Encoding.} We employ sinusoidal positional encoding \citep{transformer} to represent both timesteps and, in the case of pose-conditional view synthesis, spatial coordinates, either xy grids for the 2D case or camera rays' origins and directions for 3D, normalized to a range of $[-1, 1]$. In contrast to the original encoding scheme used to represent discrete word positions, we further scale the arguments of $\sin$ and $\cos$ by a factor of $2{\pi}s$ (with $s$ being a hyperparameter), so to increase the distinction among the positional encodings (\cref{fig_pos_enc}).

\begin{table*}[t]
\setlength{\tabcolsep}{7.7pt}
\caption{\textbf{Performance Comparison on CelebA} of classification, reconstruction, and disentanglement, considering variational, adversarial and diffusion-based approaches. \textit{Disen.} stands for Disentanglement, \textit{Comp.} for Completeness, and \textit{Info.} for Informativeness.}
\label{tab_celeba}
\centering
\scriptsize
\begin{tabular}{lccccccccc}
\rowcolor{Blue1} \textbf{Method} & \textbf{Latent Dim} & \textbf{F1 $\uparrow$} & \textbf{Disen. $\uparrow$} & \textbf{Comp. $\uparrow$} & \textbf{Info. $\uparrow$} & \textbf{PSNR $\uparrow$} & \textbf{SSIM $\uparrow$} & \textbf{FID $\downarrow$} & \textbf{LPIPS $\downarrow$} \\
\rowcolor{Blue2} \multicolumn{10}{l}{\color{DarkBlue} {Variational Approaches}} \\
Vanilla Auto-Encoder & 2048 & 66.35 & 38.94 & 29.82 & 84.52 & \textbf{21.61} & \textbf{0.906} & \textbf{65.50} & 0.327 \\
AnnealedVAE \citep{annealed_vae} & 2048 & 68.94 & 42.99 & 30.75 & 85.53 & 15.94 & 0.686 & 145.28 & 0.433 \\
FactorVAE \citep{factor_vae} & 2048 & 71.26 & 46.34 & 29.87 & 88.06 & 20.17 & 0.890 & 87.19 & 0.331 \\
DIP-VAE (I) \citep{dipvae} & 2048 & 71.94 & 49.09 & 37.58 & 89.23 & 19.83 & 0.884 & 81.02 & 0.316 \\
DIP-VAE (II) \citep{dipvae} & 2048 & 71.85 & 48.56 & 33.37 & 89.12 & 19.95 & 0.887 & 82.41 & \textbf{0.307} \\
$\beta$-VAE \citep{beta-vae} & 2048 & \textbf{71.98} & 49.04 & \textbf{32.46} & \textbf{89.37} & 19.94 & 0.886 & 78.65 & 0.314 \\
$\beta$-TCVAE \citep{tc_vae} & 2048 & 71.82 & \textbf{49.11} & 31.63 & 89.10 & 20.18 & 0.892 & 80.33 & 0.315 \\
\rowcolor{Blue2} \multicolumn{10}{l}{\color{DarkBlue} {Adversarial Approaches}} \\
VQGAN \citep{vqgan} & 256$\times$16$\times$16 & - & - & - & - & {\color{DarkBlue}\textbf{23.28*}} & 0.773* & - & \textbf{0.311*} \\
StyleGAN2,W \citep{stylegan2} & 512 & - & \textbf{53.26} & 56.15 & {\color{DarkBlue}\textbf{96.71}} & 16.76* & 0.662* & - & 0.394* \\
StyleGAN2,W+ \citep{stylegan2} & 512$\times$14 & - & 52.71 & {\color{DarkBlue}\textbf{60.03}} & 94.46 & 21.42* & \textbf{0.813*} & - & 0.345* \\
\rowcolor{Blue2} \multicolumn{10}{l}{\color{DarkBlue} {Diffusion-based Approaches}} \\
Unconditional Diffusion & - & 63.67 & 41.33 & 30.72 & 84.94 & - & - & 25.33 & - \\
DiffAE \citep{diffae} & 512 & 68.70 & 64.39 & 39.25 & 84.61 & 15.28* & 0.681* & - & 0.392* \\
DALL-E2 (with CLIP) \citep{dalle2} & 1024 & 71.08 & 51.60 & 37.82 & 87.87 & 9.34 & 0.311 & 21.91 & 0.484 \\
\rowcolor{Blue1} \textbf{SODA (ResNet50${\times}$2)} & 2048 & {\color{DarkBlue}\textbf{72.65}} & {\color{DarkBlue}\textbf{79.93}} & \textbf{53.62} & \textbf{90.44} & \textbf{18.78} & \textbf{0.859} & \textbf{9.54} & {\color{DarkBlue}\textbf{0.273}} \\
\end{tabular}
\end{table*}

\textbf{Pose Conditioning.} Throughout the paper, we experiment with several different flavors of the novel view synthesis task: either generating a view conditionally, matching a 3D pose or 2D coordinates, or alternatively, in a pose-unconditional fashion: where given a source view, the model is asked to generate arbitrary novel views at perspectives of its choice). For the conditional case, we represent each perspective by a $H{\times}W$ 2D grid -- of $(x_0,y_0)\times(x_1,y_1)$ in the 2D case, and ray positions and directions in the 3D case -- embedded by sinusoidal positional encoding and concatenated to linearly-mapped RGB channels of the corresponding view, after the first layer of the encoder and the denoiser respectively. In \cref{ablt}, we compare different ways to represent the rays, such as through normalization, by casting them on a plane or a sphere, or by summing up their positions and directions.

\textbf{Learning Rates.} For the ImageNet dataset, we maintain a different learning rate between the encoder and the denoiser, at a ratio of $\frac{\text{lr}_{\mathcal{E}}}{\text{lr}_{\mathcal{D}}}>1$. We practically implement it by following the idea of learning rate equalization \citep{progan}, scaling down the initialized weights of the encoder by a factor of $k$ (by scaling down the standard deviation of the initialization distribution), and then having the network itself scale them back up by $k$, effectively scaling the encoder's gradients by $k$. While the model is robust to the selection of the learning rate ratio, we find that a ratio of 2 yields optimal downstream results (\cref{ablt}).

\section{Datasets, Preprocessing \& Augmentations}
\label{datasets}
\subsection{Datasets Overview}
Throughout this work, we evaluate models over various datasets grouped into multiple tasks, as summarized by \cref{tab_data} and through the textual description below:

\medskip
\noindent \textbf{Representation Learning \& Reconstruction}: \\
Each image in the following datasets is associated with a category label (or for CelebA, with multiple attribute annotations). 
\begin{enumerate}[(1),leftmargin=0.8cm]
\item {\color{BlueT}Imagenet1K} \citep{imagenet}: includes diverse images of objects among 1,000 categories of e.g. animals, instruments, furniture and food items.
\item {\color{BlueT}CelebA-HQ} \citep{celeba}: features face images, annotated with 40 binary semantic properties like age, gender, or hair color; used also for quantitative disentanglement analysis.
\item {\color{BlueT}LSUN} \citep{lsun}: partitioned into multiple categories of objects (like cars, cats and horses) and scenes (e.g. bedrooms and churches); See \cref{tab_data} for full list.
\item {\color{BlueT}Animal Faces-HQ (AFHQ)} \citep{afhq}: covers various breeds of cats, dogs and wildlife.
\item {\color{BlueT}Oxford Flowers 102} \citep{flowers}: features diverse flowers from the United Kingdom.
\end{enumerate}

\medskip
\noindent \textbf{Novel View Synthesis}: \\
Each image in the following datasets is associated with the camera perspective it was captured from, expressed as a grid of ray positions and directions $\bm{r}=(\bm{o},\bm{d})$.
\begin{enumerate}[(1),leftmargin=0.8cm, resume]
\item {\color{BlueT}NMR} \citep{nmr}: consists of ShapeNet \citep{shapenet} objects' renderings at 24 fixed views, evenly spaced around a surrounding ring with constant radius and altitude; images of $64{\times}64$ resolution. We use the SoftRas data split \citep{softras}. 
\item {\color{BlueT}ShapeNet}: our custom ShapeNet renderings dataset, featuring 120 views randomly sampled from an upper hemisphere, with random azimuth $\phi$, altitude $\theta$, and radius $r \,{\in}\, [r_\text{min},r_\text{max}]$; $256{\times}256$ resolution; created by the Blender-based Kubric library \citep{kubric}.
\item {\color{BlueT}Google Scanned Objects (GSO)} \citep{gso}: includes scans of real-world household items, which we render with Blender 
following the same protocol described above.
\end{enumerate}

\medskip
\noindent \textbf{Disentanglement (Quantitative)}: \\
Each image in the following datasets is associated with discrete semantic attribute annotations.
\begin{enumerate}[(1),leftmargin=0.8cm,resume]
\item {\color{BlueT}SmallNORB} \citep{smallnorb}: contains toy images belonging to 5 categories like animals and vehicles, captured from various camera perspectives and lighting conditions.
\item {\color{BlueT}3DShapes} \citep{shapes3d}: includes images of a centered object  among varied combinations of shape, color, size and orientation (4 shapes, 8 scales, 15 orientations, and 10 possible colors for the object, wall and floor).
\item {\color{BlueT}MPI3D} \citep{mpi3d}: includes 4 splits of either synthetic or real objects, hold by a robotic arm, with different discrete attributes (4-6 shapes, 4-6 colors, 2 sizes, 3 background colors, and ${3}\times{40}\times{40}$ camera perspectives). 
\item {\color{BlueT}Caltech-UCSD Birds} (CUB-200-2011) \citep{cub}: contains images of various bird species, annotated with 312 binary semantic properties.
\end{enumerate}

\begin{figure*}[t]
\centering
\subfloat{\includegraphics[width=0.23666\linewidth]{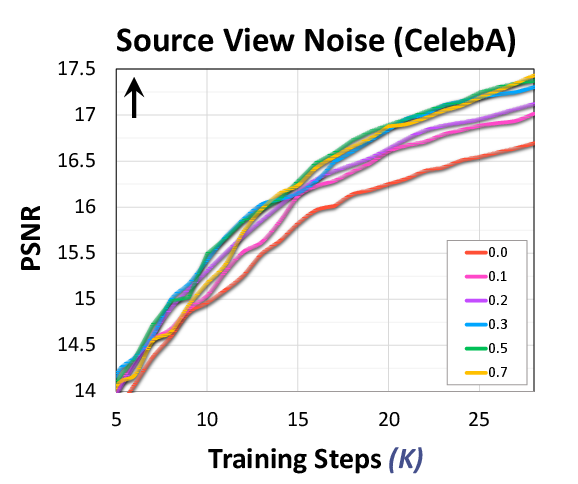}}
\hfill
\subfloat{\includegraphics[width=0.23666\linewidth]{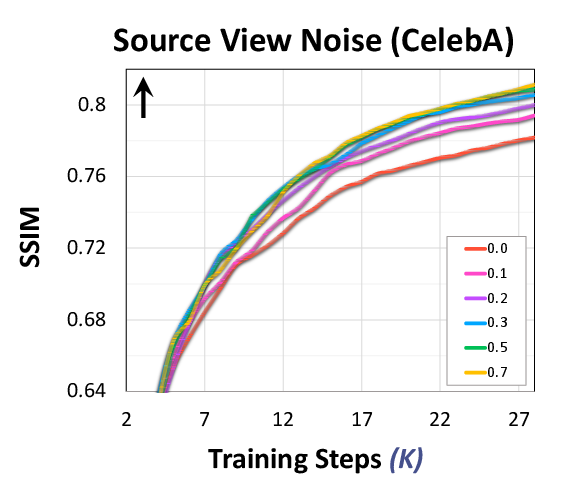}}
\hfill
\subfloat{\includegraphics[width=0.23666\linewidth]{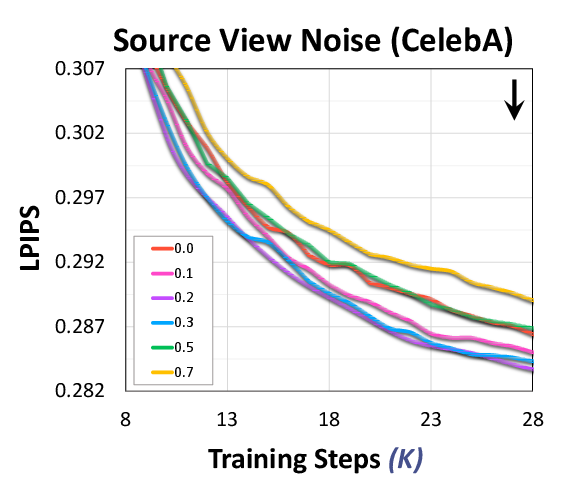}}
\hfill
\subfloat{\includegraphics[width=0.28\linewidth]{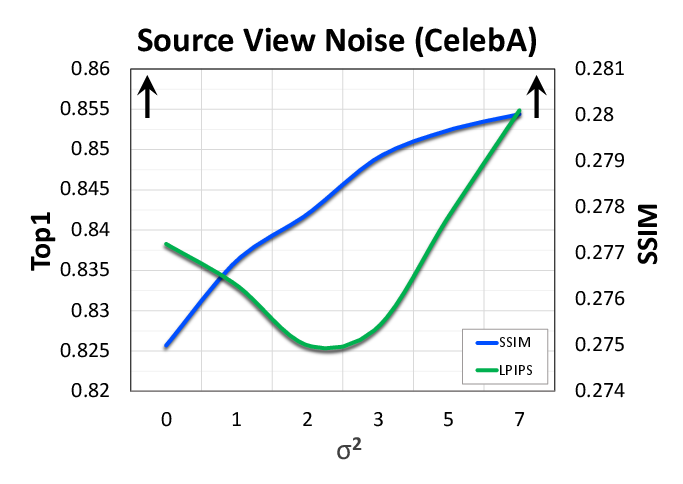}}
\caption{\textbf{Impact of source view noise} on the model's generative and representational capabilities.}
\label{fig_source_view_noise}
\end{figure*}

\subsection{Data Preprocessing}

\textbf{Resolution.} We resize all images for training and evaluation to a source resolution of $256{\times}256$, inputted into the encoder $\mathcal{E}$, and target resolution $128{\times}128$, produced by the decoder $\mathcal{D}$, with the exception of CUB and ImageNet: for the former, we center-crop and pad each image based on its associated bird's bounding box; for the latter, we first resize the target images to $256{\times}256$, and then center-crop them to $224{\times}224$, matching prior literature \citep{simclr,mae}.

We keep the model's output resolution as 128 since according to diffusion models' practices, higher-resolution images are commonly produced through cascading \citep{cascade}, where a core module first generates images of resolution 64 or 128, and these are these are subsequently post-processed by an independent super-resolution module, rather than being created as high-resolution directly. Indeed, this technique has been shown to improve the overall sample quality, and could readily fit with our approach as well.

\textbf{Normalization.} We normalize the input images fed into the encoder $\mathcal{E}$ based on ImageNet mean and variance statistics \citep{imagenet_stats}, while linearly scaling the target images of the denoising decoder $\mathcal{D}$ to the range $[-1,1]$, following the standard procedures. 

\textbf{Data Splits.} For each dataset, we either use the default splits, or if not provided, split them into 80\% training, 10\% validation and 10\% testing. Data is shuffled at training time. We note that for all the multi-view datasets: NMR, ShapeNet, GSO, and smallNorb, we intentionally keep all the views of each object exclusively grouped within one of the splits, and consequently, all the objects used for evaluation are not included in the training set. 

\subsection{Data Augmentation}

\begin{table*}[t]
\setlength{\tabcolsep}{10pt}
\caption{\textbf{Disentanglement Analysis}, comparing SODA to variational approaches on various datasets. Our model achieves improvements of 27.2-58.3\% in Disentanglement, 5.0-23.8\% in Completeness, and comparable Informativeness. Its reconstructions are often sharper and more accurate. Metrics: \textit{Disen.} stands for Disentanglement, \textit{Comp.} for Completeness, and \textit{Info.} for Informativeness. \textit{PSNR}, \textit{SSIM}, and \textit{LPIPS} respectively express pixel-wise, structural and perceptual/semantic similarity, while \textit{FID} captures sharpness and fidelity.} 
\label{tab_disen_supp}
\centering
\scriptsize
\begin{tabular}{lccccccc}
\rowcolor{Blue1} \textbf{Method} & \textbf{Disen. $\uparrow$} & \textbf{Comp. $\uparrow$} & \textbf{Info. $\uparrow$} & \textbf{PSNR $\uparrow$} & \textbf{SSIM $\uparrow$} & \textbf{FID $\downarrow$} & \textbf{LPIPS $\downarrow$} \\
\rowcolor{Blue1} \multicolumn{8}{l}{\color{BlueT} \textbf{MPI3D (Toy)}} \\
AnnealedVAE \citep{annealed_vae} & 18.43 & 21.44 & 52.47 & 29.49 & 0.852 & 118.18 & 0.169 \\
FactorVAE \citep{factor_vae} & 22.41 & 27.06 & 60.79 & 34.60 & 0.959 & 31.64 & 0.075 \\
DIP-VAE (I) \citep{dipvae} & 51.12 & 41.33 & 72.05 & 37.34 & 0.977 & 22.33 & 0.057 \\
DIP-VAE (II) \citep{dipvae} & 24.88 & 27.37 & 64.83 & 36.68 & 0.976 & 25.52 & 0.062 \\
$\beta$-VAE \citep{beta-vae} & 25.34 & 28.57 & 64.65 & 36.86 & 0.977 & 24.61 & 0.061 \\
$\beta$-TCVAE \citep{tc_vae} & 32.21 & 40.10 & 64.03 & 35.17 & 0.965 & 30.81 & 0.072 \\
\rowcolor{Blue3} SODA w/o layer mod.  & 83.44 & \textbf{55.65} & \textbf{85.38} & 50.21 & 0.998 & 2.64 & 0.015 \\
\rowcolor{Blue2} \textbf{SODA (\textit{ours})} & \textbf{87.38} & 54.79 & 84.78 & \textbf{50.72} & \textbf{0.999} & \textbf{1.49} & \textbf{0.014} \\
\rowcolor{Blue1} \multicolumn{8}{l}{\color{BlueT} \textbf{MPI3D (realistic)}} \\
AnnealedVAE \citep{annealed_vae} & 18.76 & 19.93 & 51.66 & 32.54 & 0.968 & 55.95 & 0.201 \\
FactorVAE \citep{factor_vae} & 30.80 & 36.63 & 60.82 & 34.51 & 0.980 & 29.76 & 0.181 \\
DIP-VAE (I) \citep{dipvae} & 45.61 & 42.37 & 66.59 & 36.23 & 0.986 & 24.90 & 0.169 \\
DIP-VAE (II) \citep{dipvae} & 29.63 & 35.85 & 64.85 & 36.33 & 0.986 & 25.14 & 0.170 \\
$\beta$-VAE \citep{beta-vae} & 25.14 & 26.15 & 61.48 & 36.54 & 0.987 & 24.46 & 0.168 \\
$\beta$-TCVAE \citep{tc_vae} & 35.81 & 41.07 & 66.83 & 36.71 & 0.988 & 24.41 & 0.167 \\
\rowcolor{Blue3} SODA w/o layer mod.  & 71.73 & 51.07 & 76.53 & 39.40 & 0.995 & \textbf{3.42} & 0.109 \\
\rowcolor{Blue2} \textbf{SODA (\textit{ours})} & \textbf{85.19} & \textbf{56.26} & \textbf{77.78} & \textbf{40.65} & \textbf{0.996} & 3.84 & \textbf{0.069} \\
\rowcolor{Blue1} \multicolumn{8}{l}{\color{BlueT} \textbf{MPI3D (real)}} \\
AnnealedVAE \citep{annealed_vae} & 17.62 & 17.41 & 53.60 & 31.97 & 0.963 & 40.33 & 0.083 \\
FactorVAE \citep{factor_vae} & 33.99 & 41.48 & 59.74 & 34.19 & 0.978 & 33.70 & 0.062 \\
DIP-VAE (I) \citep{dipvae} & 53.01 & 40.35 & 72.05 & 36.68 & 0.988 & 18.31 & 0.037 \\
DIP-VAE (II) \citep{dipvae} & 33.54 & 38.72 & 65.57 & 36.37 & 0.987 & 20.36 & 0.041 \\
$\beta$-VAE \citep{beta-vae} & 48.56 & 48.20 & 68.19 & 36.64 & 0.988 & 19.35 & 0.038 \\
$\beta$-TCVAE \citep{tc_vae} & 48.60 & 44.68 & 67.03 & 36.47 & 0.987 & 19.74 & 0.040 \\
\rowcolor{Blue3} SODA w/o layer mod.  & 75.47 & \textbf{51.30} & 77.32 & 41.03 & \textbf{0.998} & 1.57 & 0.007 \\
\rowcolor{Blue2} \textbf{SODA (\textit{ours})} & \textbf{81.19} & 50.92 & \textbf{77.69} & \textbf{42.51} & 0.997 & \textbf{0.61} & \textbf{0.006} \\
\rowcolor{Blue1} \multicolumn{8}{l}{\color{BlueT} \textbf{MPI3D (complex)}} \\
AnnealedVAE \citep{annealed_vae} & 21.08 & 21.27 & 57.73 & 31.37 & 0.952 & 39.71 & 0.077 \\
FactorVAE \citep{factor_vae} & 37.50 & 48.55 & 68.12 & 32.93 & 0.966 & 36.27 & 0.067 \\
DIP-VAE (I) \citep{dipvae} & 58.49 & 51.67 & 76.64 & 35.14 & 0.980 & 23.50 & 0.042 \\
DIP-VAE (II) \citep{dipvae} & 35.00 & 40.21 & 72.30 & 35.03 & 0.979 & 27.44 & 0.045 \\
$\beta$-VAE \citep{beta-vae} & 51.82 & 50.70 & 74.47 & 35.58 & 0.982 & 22.78 & 0.039 \\
$\beta$-TCVAE \citep{tc_vae} & 42.99 & 46.39 & 74.26 & 35.42 & 0.981 & 24.90 & 0.042 \\
\rowcolor{Blue3} SODA w/o layer mod.  & 86.55 & 56.60 & 77.43 & \textbf{44.23} & \textbf{0.998} & 0.47 & 0.007 \\
\rowcolor{Blue2} \textbf{SODA (\textit{ours})} & \textbf{89.76} & \textbf{56.98} & \textbf{78.22} & 43.18 & 0.997 & \textbf{0.44} & \textbf{0.006} \\
\rowcolor{Blue1} \multicolumn{8}{l}{\color{BlueT} \textbf{3DShapes}} \\
AnnealedVAE \citep{annealed_vae} & 58.14 & 66.97 & 91.23 & 30.77 & 0.994 & 54.83 & 0.063 \\
FactorVAE \citep{factor_vae} & 87.62 & \textbf{87.52} & 98.64 & 30.37 & 0.994 & 48.89 & 0.058 \\
DIP-VAE (I) \citep{dipvae} & 99.75 & 82.59 & 99.97 & 34.23 & 0.997 & 31.04 & 0.031 \\
DIP-VAE (II) \citep{dipvae} & 99.22 & 83.01 & \textbf{99.99} & 33.87 & 0.997 & 32.41 & 0.033 \\
$\beta$-VAE \citep{beta-vae} & \textbf{99.87} & 83.51 & 99.96 & 33.93 & 0.997 & 29.52 & 0.032 \\
$\beta$-TCVAE \citep{tc_vae} & 90.92 & 74.78 & 99.81 & 34.62 & 0.998 & 29.03 & 0.031 \\
\rowcolor{Blue3} SODA w/o layer mod.  & 92.18 & 78.21 & 99.27 & 51.75 & 0.9997 & 0.36 & {0.0003} \\
\rowcolor{Blue2} \textbf{SODA (\textit{ours})} & 98.60 & 84.76 & 98.88 & \textbf{52.54} & \textbf{0.9999} & \textbf{0.32} & \textbf{0.0002} \\
\rowcolor{Blue1} \multicolumn{8}{l}{\color{BlueT} \textbf{SmallNORB}} \\
AnnealedVAE \citep{annealed_vae} & 21.50 & 17.62 & 60.30 & 28.40 & 0.900 & 164.94 & 0.271 \\
FactorVAE \citep{factor_vae} & 37.01 & 44.58 & 68.52 & 27.67 & 0.885 & 148.40 & 0.267 \\
DIP-VAE (I) \citep{dipvae} & 34.13 & 35.51 & \textbf{71.85} & 28.92 & 0.907 & 114.48 & 0.226 \\
DIP-VAE (II) \citep{dipvae} & 37.94 & 45.76 & 69.89 & 29.15 & 0.909 & 122.10 & 0.232 \\
$\beta$-VAE \citep{beta-vae} & 37.79 & 42.40 & 71.01 & \textbf{29.25} & \textbf{0.912} & 120.31 & \textbf{0.229} \\
$\beta$-TCVAE \citep{tc_vae} & 37.43 & 43.51 & 70.75 & 29.20 & 0.909 & 120.28 & 0.230 \\
\rowcolor{Blue3} SODA w/o layer mod.  & 63.12 & 45.82 & 68.36 & 16.06 & 0.756 & 47.90 & 0.253 \\
\rowcolor{Blue2} \textbf{SODA (\textit{ours})} & \textbf{72.60} & \textbf{51.56} & 70.19 & 15.47 & 0.734 & \textbf{44.81} & 0.235 \\
\rowcolor{Blue1} \multicolumn{8}{l}{\color{BlueT} \textbf{CUB}} \\
AnnealedVAE \citep{annealed_vae} & 37.53 & 11.59 & 91.71 & 14.73 & 0.244 & 275.35 & 0.716 \\
FactorVAE \citep{factor_vae} & 39.14 & 11.68 & 92.04 & \textbf{16.26} & \textbf{0.517} & 244.70 & 0.637 \\
DIP-VAE (I) \citep{dipvae} & 36.44 & 10.61 & \textbf{92.12} & 15.05 & 0.421 & 233.65 & 0.642 \\
DIP-VAE (II) \citep{dipvae} & 37.43 & 12.19 & 92.05 & 14.97 & 0.409 & 227.14 & 0.647 \\
$\beta$-VAE \citep{beta-vae} & 38.22 & 11.88 & 92.06 & 15.10 & 0.428 & 234.52 & 0.639 \\
$\beta$-TCVAE \citep{tc_vae} & 40.43 & 13.39 & 91.95 & 15.26 & 0.389 & 237.96 & 0.652 \\
\rowcolor{Blue3} SODA w/o layer mod.  & 62.05 & 14.35 & 87.86 & 12.96 & 0.423 & 20.30 & 0.503 \\
\rowcolor{Blue2} \textbf{SODA (\textit{ours})} & \textbf{65.40} & \textbf{17.43} & 88.10 & 13.04 & 0.344 & \textbf{17.21} & \textbf{0.492} \\
\end{tabular}
\vspace{50pt}
\end{table*}

We study several augmentation schemes, applied for different tasks and datasets: by default, we use random resize cropping, horizontal flipping and optionally RandAugment \citep{rand_augment} data augmentation on both the source and target views $\bm{x'}$ and $\bm{x}$ (encoded and denoised respectively). Specifically, at every training step, we randomly augment each view, at the rates specified in \cref{tab_impl}. To train the subsequent downstream classifier, we perform cropping and flipping only, and finally, at evaluation time, perform only center-cropping, following the standard linear probing protocols of prior self-supervision learning works \citep{dino, mae}. When training the diffusion model, we also find it conducive to add low Gaussian noise to the encoded source view, similarly to the noise added to the denoised target view.

Meanwhile, for multi-view 3D datasets such as NMR, GSO and ShapeNet, we do not apply data augmentations, and instead, randomly sample one view as the source and another as the target, further supplied by their respective camera perspectives (\cref{supp_impl}). In this case, we allow for conditioning on multiple source views, and conduct experiments over $k\,{\in}\,[1-9]$ sources. Lastly, to illustrate the ability of SODA to learn useful representations even without relying on data augmentation, we perform ablations on datasets used as is, forgoing augmentations of any kind. 

\section{Evaluation \& Metrics}
\label{eval}

\begin{figure*}[t]
\centering
\subfloat{\includegraphics[width=0.2\linewidth]{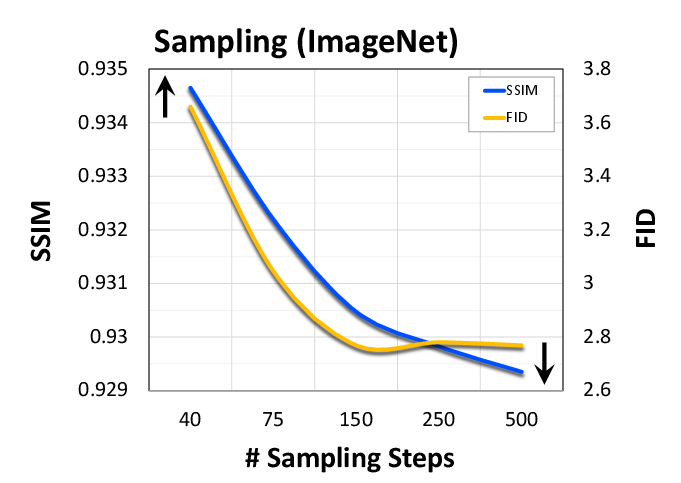}}
\hfill
\subfloat{\includegraphics[width=0.2\linewidth]{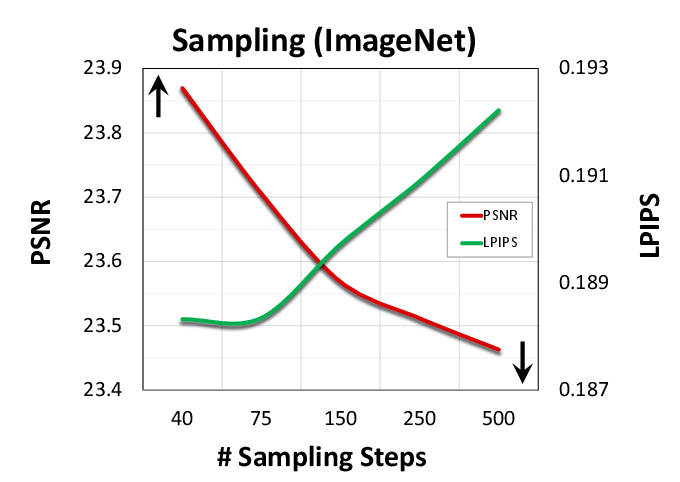}}
\hfill
\subfloat{\includegraphics[width=0.2\linewidth]{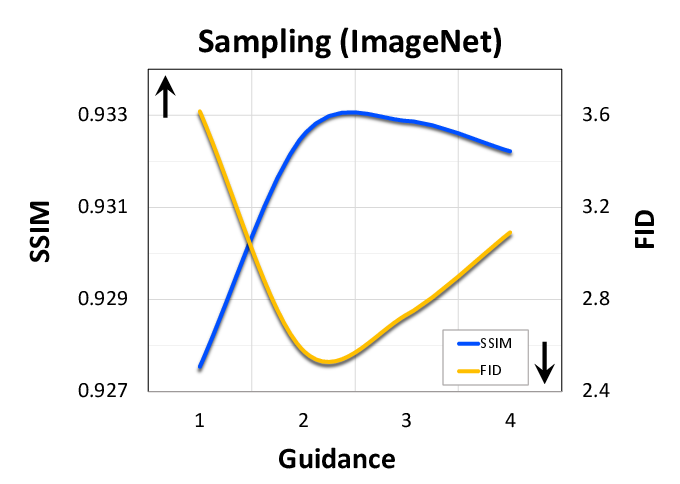}}
\hfill
\subfloat{\includegraphics[width=0.2\linewidth]{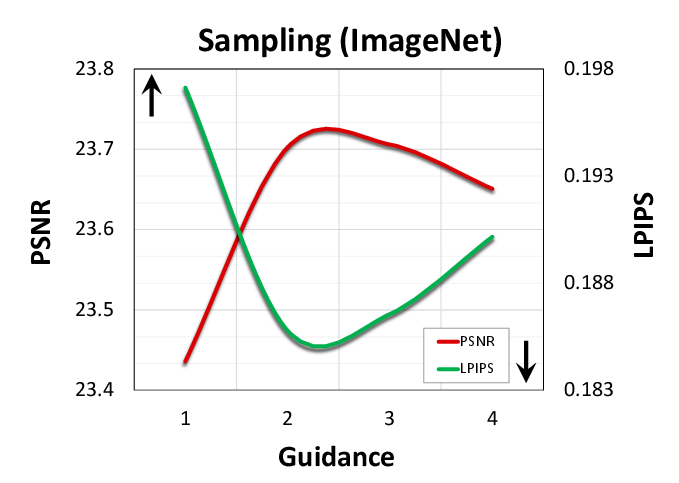}}
\hfill
\subfloat{\includegraphics[width=0.2\linewidth]{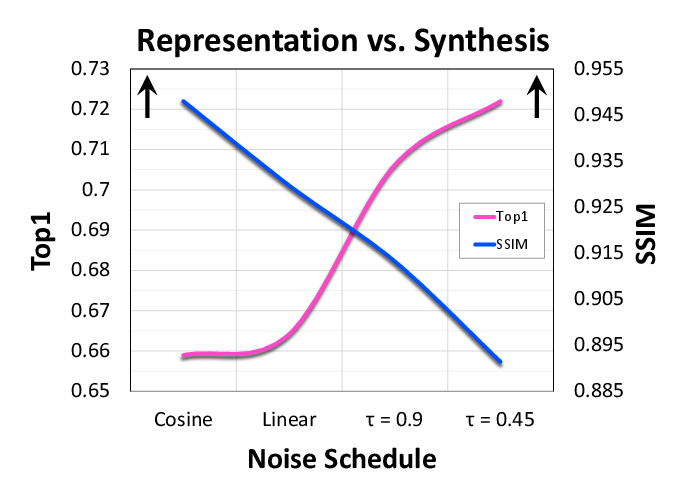}}

\subfloat{\includegraphics[width=0.2\linewidth]{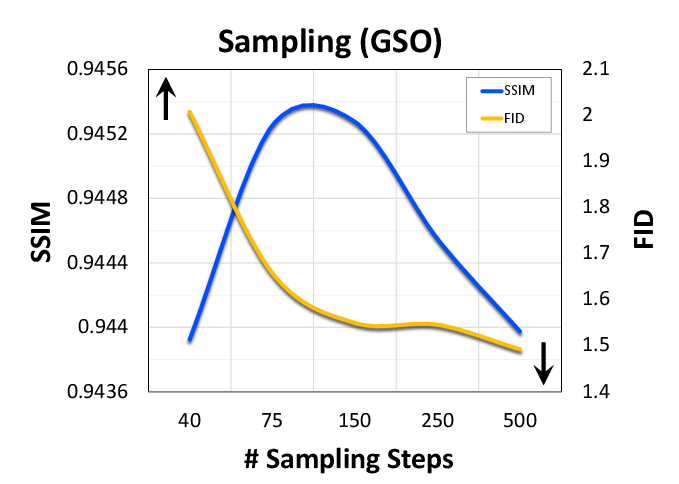}}
\hfill
\subfloat{\includegraphics[width=0.2\linewidth]{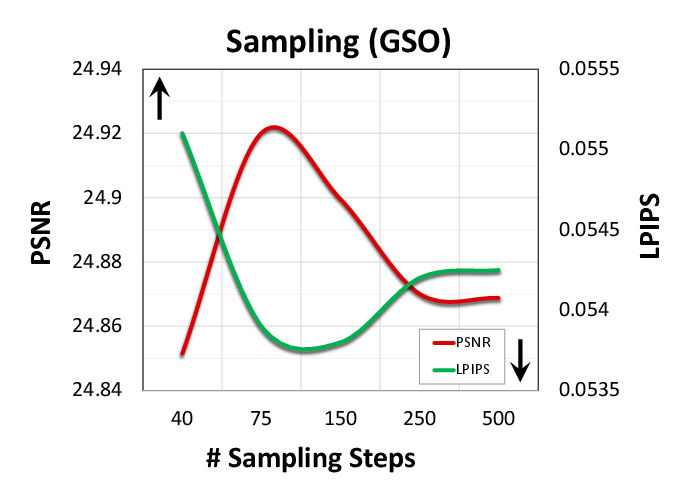}}
\hfill
\subfloat{\includegraphics[width=0.2\linewidth]{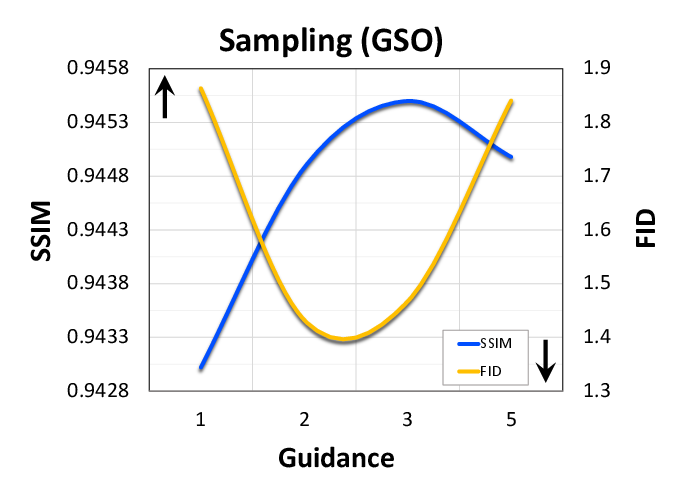}}
\hfill
\subfloat{\includegraphics[width=0.2\linewidth]{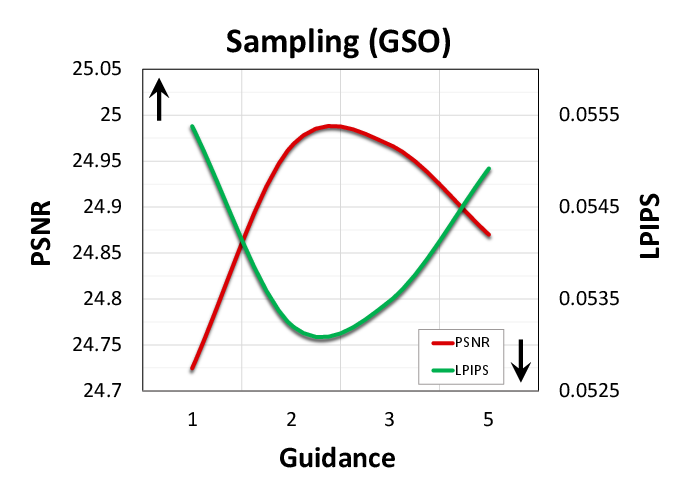}}
\hfill
\subfloat{\includegraphics[width=0.2\linewidth]{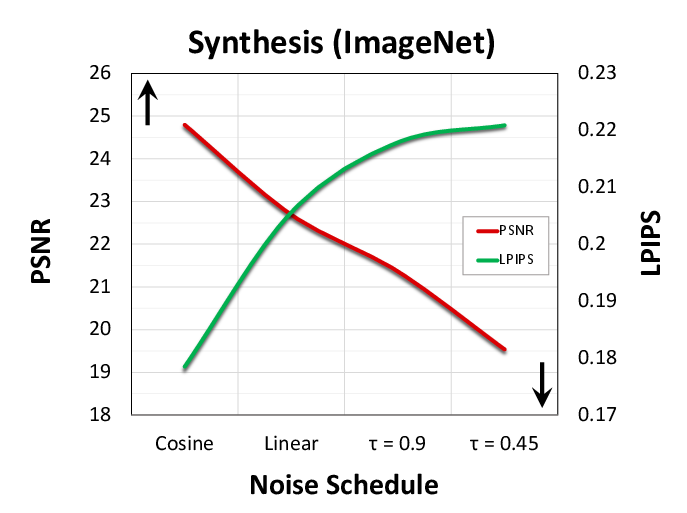}}
\caption{\textbf{Impact of classifier-free guidance and number of sampling steps} and noise schedule on the model's performance.}
\label{fig_guidance_sampling}
\end{figure*}

We explore SODA for multiple types of tasks and purposes: downstream linear-probe classification and disentanglement analysis for assessing the quality of the learned representations, as well as image reconstruction and novel view synthesis for evaluating the model's generative capabilities. These skills are measured both through qualitative inspection of the latent space, with visualizations that demonstrate its impact on the model's outputs (including in particular latent interpolations and unsupervised attribute discovery), as well as through an assortment of metrics that quantify each of the capabilities as discussed below.

\subsection{Linear Probing}

In \cref{cls}, we analyze the model's learned latent representations by measuring their predictive performance on a downstream classification task. Following the common evaluation protocol \citep{dino, mae, simclr}, we first train our model on a collection of images, and then fit a linear classifier that considers the latent encodings produced by the model and use them to predict each respective image's category or semantic attributes. The classifier is either trained on the frozen representations $\bm{z}$ subsequently to the training of the diffusion model, or alternatively, trained with it in tandem by blocking the gradient flow between the two networks -- we find that both approaches achieve similar results. 

When training the classifier, we refrain from applying weight decay, and adhere to either light augmentation of cropping and flipping for ImageNet or no augmentation in other cases. The latents $\bm{z}$ are normalized before being fed to the classifier, concretely, by processing them with an un-parameterized batch normalization \citep{mae}, which only tracks mean and variance statistics and lacks the follow-up affine transformation. After normalizing the latents, we use 0.1 dropout for regularization, and for ImageNet, apply label smoothing of 0.1. Since the annotated datasets we explore all have discrete labels, we use softmax cross entropy to train the classifier, and report its performance along metrics such as F1 for binary attributes, and top1 accuracy for other ones.

\subsection{Image synthesis}
\label{metrics_synth}
In \cref{recon}, we analyze the capacity of SODA to both reconstruct an input view and generate novel views. Given a target image $x$, we assess the quality of a synthesized output $\hat{x}$ through multiple complementary metrics that range from visual to semantic similarity: 

\begin{enumerate}[(1),leftmargin=0.8cm]
\item \textbf{Peak Signal-too-Noise Ratio (PSNR)} $\uparrow$ \textit{(measured in dB)} \citep{psnr}: is directly derived from the mean MSE between $x$ and $\hat{x}$, and it thereby measures \textbf{pixel-wise} similarity. It may rate a blurry estimation as highly consistent with the target, as long as they match well with each other \textit{on average}.

\item \textbf{Structural Similarity Index Measure (SSIM)} $\uparrow$ \textit{(ranges between $[-1,1]$)} \citep{ssim}: compares images along three perceptual factors: luminance, contrast and \textbf{structure}, and is thus better correlated with the Human Visual System (HVS).

\item \textbf{Learned Perceptual Image Patch Similarity (LPIPS)} $\downarrow$\textit{(often normalized to be in $[0,1]$)} \citep{lpips}: computes the distance between the target and synthesized images in the feature space of a supervised pre-trained network, such as VGG \citep{vgg}, and therefore serves as an indicator for \textbf{semantic} similarity.

\item \textbf{Fréchet inception distance (FID) score} $\downarrow$ \textit{(is ${\geq}0$)} \citep{fid}: quantifies \textbf{realism and sharpness} of the generated images by comparing their distribution to that of the target ground-truth images. It concretely achieves it by considering the mean and variance of each, in a latent feature space, e.g. of the Inception model \citep{inception}. When assessing unconditionally-generated images, the FID score further expresses their diversity, but in the case of conditional synthesis, either as reconstructions or with pose conditioning, it mainly reflects their fidelity, sharpness and lack of distortions (also known as R-FID in this context). 
\end{enumerate}

\noindent For fair comparison, we compute these metrics over all approaches using the same metrics' implementations, and specifically, casting the images to the range of $[0, 1]$ for PSNR, $[-1, 1]$ for FID and LPIPS, and using a uniform kernel to calculate SSIM scores.

\subsection{Disentanglement}
\label{metrics_disen}

In \cref{disen}, we examine the latent space of our model and assess its degree of disentanglement and controllability through quantitative and qualitative evaluation methods:

\textbf{DCI metrics (Disentanglement, Completeness \& Informativeness)} $\uparrow$ \textit{(at a range of $[0,100\%]$)} \citep{dci}: measures the 1:1 alignment between  the latent representation $\bm{z}$ and the natural (ground-truth) factors of variation $\bm{c}$. \textbf{Disentanglement} reflects the extant to which each latent variable $\hat{z}_j$ (the $j$'th axis of the vector $\bm{z}$) corresponds to a unique natural factor $c_j$. \textbf{Completeness} inversely measures the extant to which each natural factor $c_j$ is captured by a single latent variable $\hat{z}_j$. Finally, \textbf{Informativeness} indicates the predictability of the natural factors $c$ from the latent encoding $\bm{z}$. These metrics are derived from the normalized importance matrix of a learned classifier and its performance, where the classifier is based on either gradient boosting or Lasso (we use the former). Our implementation of these metrics closely follows Locatello \etal \citep{disen_lib}. 
 
\textbf{Latent Interpolation}: we randomly pick two images $\bm{x}_1,\bm{x}_2$ from each dataset, encode them to obtain $\bm{z}_1,\bm{z}_2$, and then decode back the latents along a linear segment that connects between the endpoints: $\bm{z}_1 + (\bm{z}_2-\bm{z}_1){\cdot}t$ for $t\,{\in}\,[0,1]$, which results in a visualization of the latent traversal.

\textbf{Principal Component Analysis (PCA)} \citep{pca, pca_on_stylegan}: we encode a sample set of $N$ images (1,000-10,000), and perform PCA decomposition over the obtained latents $\bm{z}_{i=1}^N$, which yields the latent directions $\bm{s}_j$ of the greatest variation. We then traverse the latent space along the discovered directions: $\bm{z} + \bm{s}_{j}t$ for $t\,{\in}\,[-\sqrt{\lambda_j},\sqrt{\lambda_j}]$ where $\lambda_j$ is the respective eigenvalue and $\sqrt{\lambda_j}$ is the standard deviation along direction $\bm{s}_j$. Doing so allows us to visualize the impact of these latent directions on the model's generated images, and indeed, we find they strongly correlate with semantically-meaningful manipulations.

Thanks to layer modulation (\cref{masking}), we can further perform PCA over chosen sub-vectors of $\bm{z}$ that are responsible for guiding decoder's layers of interest. This enables the discovery of latent directions that control particular levels of granularity, from low-frequency structural aspects to high-frequency factors like texture and color, enhancing the model's overall controllability.

\begin{table}[t]
\caption{\textbf{Ablations on ShapeNet}, varying the \textbf{camera viewpoint encoding schemes}, including \textbf{coordinate} system: Polar or Cartesian, rays' representation \textbf{method}: by origin and direction $[\bm{o},\bm{d}]$, or with a weighted sum $\bm{o}\,+\,s_d{\cdot}\bm{d}$, and \textbf{conditioning} mode: through 2D grid concatenation or vector-based modulation (either of \textit{absolute} or \textit{relative} camera perspective).}
\label{tab_pos_ablt}
\centering
\scriptsize
\begin{tabular}{lllcccc}
\rowcolor{Blue1} \textbf{Method} & \textbf{Coordinates} & \textbf{PSNR $\uparrow$} & \textbf{SSIM $\uparrow$} & \textbf{FID $\downarrow$} & \textbf{LPIPS $\downarrow$} \\
Concat & Cartesian & 27.21 & 0.945 & 0.82 & 0.040 \\
\rowcolor{Blue2} Concat & Polar & 27.10 & 0.940 & 0.93 & 0.041 \\
Sphere & Cartesian & 27.12 & 0.946 & 0.75 & 0.040 \\
\rowcolor{Blue2} Normalized & Cartesian & 27.00 & 0.941 & 0.78 & 0.041 \\
Plane & Cartesian & 27.03 & 0.943 & 0.78 & 0.041 \\
\rowcolor{Blue1} \textbf{Sphere} & \textbf{Polar} & \textbf{27.42} & \textbf{0.947} & \textbf{0.74} & \textbf{0.039} \\
\\
\rowcolor{Blue1} \textbf{Conditioning} & \textbf{Coordinates} & \textbf{PSNR $\uparrow$} & \textbf{SSIM $\uparrow$} & \textbf{FID $\downarrow$} & \textbf{LPIPS $\downarrow$} \\
Absolute & Polar & 23.81 & 0.870 & 2.65 & 0.059 \\
\rowcolor{Blue2} Grid+Relative & Polar & 26.81 & 0.941 & 0.87 & 0.040 \\
Grid+Absolute & Polar & 27.27 & 0.946 & 0.80 & 0.040 \\
\rowcolor{Blue1} \textbf{2D Grid} & \textbf{Polar} & \textbf{27.42} & \textbf{0.947} & \textbf{0.74} & \textbf{0.039} \\
\end{tabular}
\end{table}

\begin{table}[t]
\caption{\textbf{Ablations on ImageNet} for feature modulation (\textit{mod.}), evaluated through reconstruction and classification.}
\label{tab_ablt_imagenet}
\centering
\scriptsize
\begin{tabular}{lccccc}
\rowcolor{Blue1} \textbf{Ablation} & \textbf{PSNR $\uparrow$} & \textbf{SSIM $\uparrow$} & \textbf{FID $\downarrow$} & \textbf{LPIPS $\downarrow$} & \textbf{Top1 $\uparrow$}\\
\rowcolor{Blue2} w/o scale mod. & 21.33 & 0.910 & 5.33 & 0.224 & 70.31 \\
w/o layer mod. & 17.61 & 0.800 & 9.99 & 0.377 & 68.10 \\
\rowcolor{Blue2} sum mod. & 13.78 & 0.541 & 15.24 & 0.483 & 61.46 \\
concat mod. & 13.57 & 0.522 & 16.68 & 0.494 & 59.87 \\
\rowcolor{Blue1} \textbf{SODA (\textit{default})} & \textbf{23.63} & \textbf{0.931} & \textbf{2.77} & \textbf{0.191} & \textbf{72.24} \\
\end{tabular}
\end{table}

\textbf{Classifier-based Attribute Manipulation}: For datasets with binary attributes annotations, such as CelebA and CUB, we can produce similar visualizations to the ones described above by examining the weight matrix's rows of the linear probes we train for the downstream classification experiments (\cref{cls}). Indeed, these probes are trained to capture the latent directions that correspond to the presence or absence of the semantic attribute annotations that accompany the datasets we study. The key difference between the PCA-based approach and this technique is that the former is unsupervised while the latter is not.

\section{Baselines}
\label{baselines} 
For each of the tasks we explore, we compare our model to the respective leading approaches, as well as to additional ablated baselines that we design. Here, we list and review all the baseline methods we compare to.

\begin{figure}[t]
\centering
\scriptsize
  \includegraphics[width=0.32\linewidth]{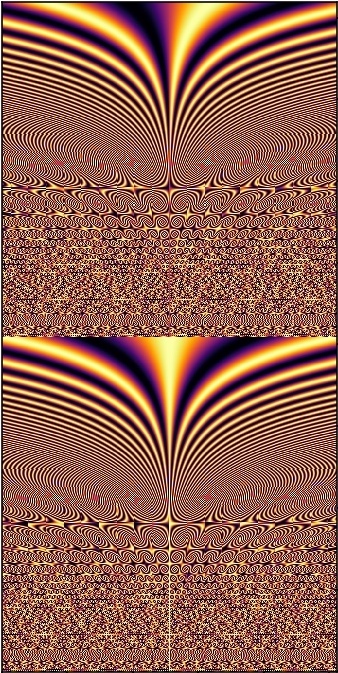}
  \includegraphics[width=0.32\linewidth]{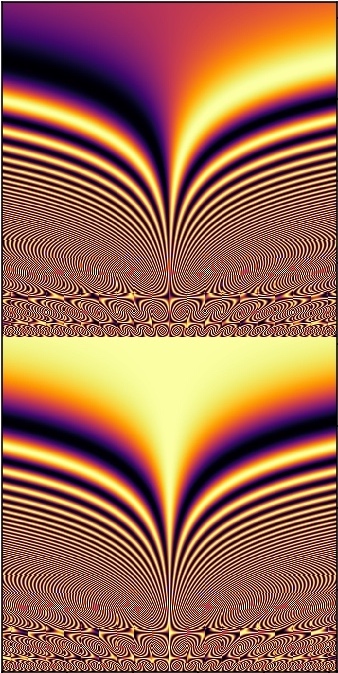}
  \includegraphics[width=0.32\linewidth]{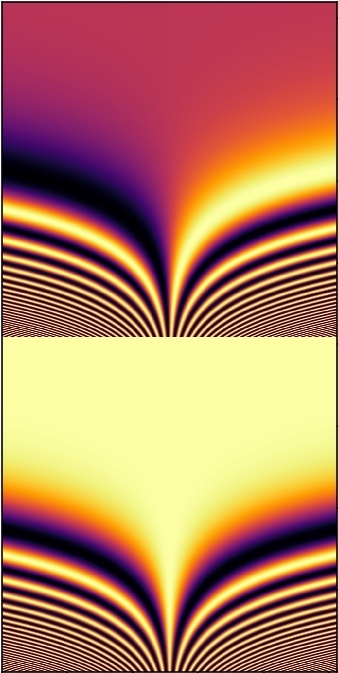}
  \caption{\textbf{Positional Encoding Scales.} We visualize the positional encodings in the range of [-1,1], with each embedding visualized vertically within each plot. When the positional encoding scale is set to a too low value (\textbf{right column}), the encodings are less distinctive and their capacity is under-utilized (as many of the features (rows) get the the same value for all positions). Meanwhile, a too high value (\textbf{left column}) damages the positional encodings empirical performance. We thus make sure to select a medium scale (\textbf{middle column}) for optimal performance.}
\label{fig_pos_enc}
\end{figure}

\subsection{Task-Agnostic Baselines}
First, we implement multiple baselines and ablated models within our diffusion codebase, and report their performance across the range of tasks: 

\begin{enumerate}[(1),leftmargin=0.8cm]
\item {\color{BlueT} EncDec}: a vanilla encoder-decoder $\bm{\hat{x}}=\mathcal{D}(\mathcal{E}(\bm{x'}))$, sharing the same encoder and decoder architectures as SODA (for $\mathcal{D}$, the decoding module of the UNet), but being trained to generate the target image $\bm{x}$ from scratch, with no denoising. We train this model both with and without ($\bm{x}=\bm{x'}$) data augmentation.

\begin{table*}[t]
\caption{\textbf{Datasets Configuration \& Statistics}. ($\star$) LSUN sizes per partition: Bedrooms (3.03M), Church Outdoors (126K), Bird (2.31M), Car (5.52M), Cat (1.66M), Dog (5.05M), Horse (2.00M).} 
\label{tab_data}
\centering
\scriptsize
\setlength{\tabcolsep}{6.5pt}
\begin{tabular}{lcccccccccc}
\rowcolor{Blue1} \textbf{Dataset} & \textbf{Size} & \textbf{Resolution} & \textbf{\# Categories} & \textbf{Augmentation} & \textbf{Source View}  & \textbf{ResNet} & \textbf{Learning} & \textbf{Batch} & \textbf{Guidance} \\
\rowcolor{Blue1} & & \textbf{(Raw)} & \textbf{/ Attributes} & \textbf{/ Views} & \textbf{Noise Scale} & \textbf{Size} & \textbf{Rate} & \textbf{Size} & \\
\rowcolor{Blue2} Imagenet1K \citep{imagenet} & 1.28M & Varied & 1000 & RandAugment & 0.10 & 50${\times}2$ & $4{\times}10^{-4}$ & 4096 & 2 \\
CelebA-HQ \citep{celeba} & 30K & 1000 & 2${\times}$40 & Gaussian Noise & 0.22 & 50${\times}2$ & $4{\times}10^{-4}$ & 4096 & 2 \\
\rowcolor{Blue2} LSUN \citep{lsun} & ($\star$) & Varied & - & Gaussian Noise & 0.22 & 50${\times}2$& $4{\times}10^{-4}$ & 4096 & 2 \\
AFHQ \citep{afhq} & 15K & 512 & 3 & Gaussian Noise & 0.22 & 50${\times}2$& $4{\times}10^{-4}$ & 4096 & 2 \\
\rowcolor{Blue2} NMR \citep{nmr} & 1.05M & 64 & 13 & 24$\times$43.8K & 0.00 & 50 & $2{\times}10^{-4}$ & 2048 & 2 \\
ShapeNet \citep{shapenet} & 6.28M & 256 & 55 & 120$\times$52K & 0.00 & 50 & $2{\times}10^{-4}$ & 2048 & 2 \\
\rowcolor{Blue2} GSO \citep{gso} & 120K & 256 & 17 & 120$\times$1K & 0.00 & 50 & $2{\times}10^{-4}$ & 2048 & 2.5 \\
SmallNORB \citep{smallnorb} & 24.3K & 96 & 18,5,9,6 & Gaussian Noise & 0.05 & 18 & $1{\times}10^{-4}$ & 1024 & 3 \\
\rowcolor{Blue2} 3D-Shapes \citep{cub} & 480K & 64 & 4,10,10,10,15,8 & Gaussian Noise & 0.05 & 18 & $1{\times}10^{-4}$ & 1024 & 3 \\
MPI3D (Toy) \citep{mpi3d} & 1.03M & 64 & 6,6,2,3,3,40,40 & Gaussian Noise & 0.05 & 18 & $1{\times}10^{-4}$ & 1024 & 3  \\
\rowcolor{Blue2} MPI3D (Realistic) \citep{mpi3d} & 1.03M & 64 & 6,6,2,3,3,40,40 & Gaussian Noise & 0.05 & 18 & $1{\times}10^{-4}$ & 1024 & 3  \\
MPI3D (Real) \citep{mpi3d} & 1.03M & 64 & 6,6,2,3,3,40,40 & Gaussian Noise & 0.05 & 18 & $1{\times}10^{-4}$ & 1024 & 3  \\
\rowcolor{Blue2} MPI3D (Complex) \citep{mpi3d} & 461K & 64 & 4,4,2,3,3,40,40 & Gaussian Noise & 0.05 & 18 & $1{\times}10^{-4}$ & 1024 & 3  \\
Caltech-UCSD Birds (CUB) \citep{cub} & 11.8K & Varied & 2${\times}$312 & Gaussian Noise & 0.05 & 18 & $1{\times}10^{-4}$ & 1024 & 3 \\
\rowcolor{Blue2} Oxford Flowers \citep{flowers} & 8.2K & Varied & 102 & Gaussian Noise & 0.05 & 18 & $1{\times}10^{-4}$ & 1024 & 3 \\
\end{tabular}
\end{table*}

\item {\color{BlueT} Uncond}: an unconditional diffusion model $\mathcal{D}$ (see also  DDAE \citep{ddae}). To obtain an encoding $\bm{z}$ for an image $\bm{x}$, we compute $\mathcal{D}(\bm{\tilde{x}})$ over a lightly-noised version of $\bm{x}$ and pool the activations from the middle layer of the UNet denoiser (The layer index and noise degree are hyperparameters chosen for highest performance).

\item {\color{BlueT} Palette}: a diffusion model that instead of having a dedicated encoder $\mathcal{E}$, concatenates the source image $\bm{x'}$ to the denoised image $\bm{x}_t$ directly, and inputs both of them to a UNet denoiser $\mathcal{D}(\bm{x}_t,\bm{x'})$ (also known as Image-to-Image diffusion model \citep{palette}).

\item {\color{BlueT} unCLIP (Dall-E2)} \citep{dalle2}: a diffusion model that relies on a frozen pretrained CLIP \citep{clip} as the encoder $\mathcal{E}$. We emphasize that we do not refer here to the already trained Dall-E2 model, but rather to its architecture, and so we train its denoiser (in a comparable size to our model) from scratch along with the frozen pre-trained CLIP encoder, for each dataset of interest.

\item {\color{BlueT} w/o bottleneck}: an ablation of SODA with no bottleneck, which rather encodes the input image $\bm{x'}$ into a 2D feature grid $\bm{z}^{w{\times}h}$, with no global pooling, and conditions the denoising on it through cross-attention (similarly to text-to-image diffusion models \citep{glide}).

\item {\color{BlueT} w/o modulation}: an ablation of SODA that broadcasts and concatenates the latent $\bm{z}$ to linearly-mapped RGB channels of the denoised image $\bm{x_t}$, instead of applying modulation through adaptive group normalization (also called a spatial broadcast decoder \citep{spatial_broadcaster}).
\end{enumerate}

\subsection{Linear-Probe Classification}
For downstream classification, we compare our model to a diverse array of leading self-supervised learning approaches: \textbf{generative} methods like {\color{BlueT}MAE} \citep{mae}, {\color{BlueT}BEIT} \citep{beit} and {\color{BlueT}iGPT} \citep{igpt} split each image into a grid of tokens or patches, mask some patches and predict them back from the unmasked ones, oftentimes using a transformer backbone.

Meanwhile, \textbf{discriminative} approaches leverage contrastive learning (as in {\color{BlueT}SimCLR} \citep{simclr}), clustering techniques (as in {\color{BlueT}SwAV} \citep{swav}), and distillation (as in {\color{BlueT}DINO} \citep{dino} and {\color{BlueT}BYOL} \citep{byol}) to derive visual representations. At the core of these methods is a strong reliance on rich data augmentations, which are essentially the driving force that allows the to perform unsupervised clustering.

Consequently, contrastive learning approaches operate well at tasks that involve identification of an image's category, as is the case for ImageNet, but may struggle to capture finer traits that are altered by the augmentations. The semantic properties they may or may not encode into the learned representations heavily depend on the particularities of the data augmentation scheme they employ, since they are basically encouraged to form a latent space that is invariant to the augmentation applied, instead casting different augmentations into similar representations.

Contrary to these two kinds of approaches, both of which are unsuitable for high-quality image generation, SODA stands out being able to both encode input images into meaningful latents, and also \textbf{synthesize} back crisp output images, conditionally and unconditionally. It learns \textbf{compact} and disentangled representations, which contrast with the large, potentially discrete, 2D grids learned by alternative approaches, and as demonstrated in \cref{cls}, is \textbf{robust to the chosen data augmentation} scheme, operating well even in its absence.

Our comparison to the approaches discussed in this subsection relies on the performance reports in their respective publications over the ImageNet1K dataset, with the exception of the crop+flip accuracy for SwAV and DINO's for which we retrain the models.

\begin{table*}[t]
\caption{\textbf{Ablations on CelebA}, evaluted through classification, reconstruction, and disentanglement. The no-bottleneck ablation ($\star$) has skip connections between the model's input and output, making the reconstruction task trivial, but simultaously damaging the learned representations' quality. \textit{Disen.} stands for Disentanglement, \textit{Comp.} for Completeness, and \textit{Info.} for Informativeness.} 
\label{tab_celeba_ablt}
\centering
\scriptsize
\begin{tabular}{lcccccccc}
\rowcolor{Blue1} \textbf{Ablation} & \textbf{F1 $\uparrow$} & \textbf{Disen. $\uparrow$} & \textbf{Comp. $\uparrow$} & \textbf{Info. $\uparrow$} & \textbf{PSNR $\uparrow$} & \textbf{SSIM $\uparrow$} & \textbf{FID $\downarrow$} & \textbf{LPIPS $\downarrow$} \\
w/o bottleneck & 58.89 & 36.61 & 26.12 & 83.15 & {\color{Gray} 27.29$^\star$} & {\color{Gray} 0.985$^\star$} & {\color{Gray} 6.61$^\star$} & {\color{Gray} 0.099$^\star$} \\
\rowcolor{Blue2} w/o modulation & 60.38 & 64.51 & 40.85 & 84.11 & 14.53 & 0.734 & 20.58 & 0.332 \\
w/o layer modulation & 70.90 & 67.25 & 42.53 & 87.06 & 16.79 & 0.821 & 10.35 & 0.288 \\
\rowcolor{Blue2} w/o layer mask & 70.36 & 68.41 & 43.41 & 87.45 & 16.98 & 0.820 & 11.39 & 0.285 \\
w/o scale modulation & 69.65 & 67.41 & 44.84 & 87.63 & 16.53 & 0.816 & 10.49 & 0.290 \\
\rowcolor{Blue2} sum modulation & 68.31 & 61.12 & 40.53 & 85.88 & 16.71 & 0.816 & 17.95 & 0.290 \\
concat modulation & 67.58 & 61.34 & 40.67 & 86.02 & 16.67 & 0.816 & 17.35 & 0.289 \\
\rowcolor{Blue1} \textbf{SODA (ResNet50, \textit{default})} & \textbf{71.63} & \textbf{73.90} & \textbf{48.81} & \textbf{87.64} & \textbf{18.24} & \textbf{0.842} & \textbf{10.09} & \textbf{0.275} \\
\end{tabular}
\end{table*}

\subsection{Image Reconstruction}
We examine the performance of varied models for the task of image reconstruction: {\color{BlueT}Dall-E} \citep{dalle} and {\color{BlueT}VQGAN} \citep{vqgan} employ a \textbf{discrete variational} auto-encoder \citep{vqvae}, which casts input images into 2D token grids, based on a trainable codebook. These approaches then couple the auto-encoder with a prior-distribution model, to enable unconditional image synthesis. However, for our purposes (image reconstruction), we consider the auto-encoder module only.

The \textbf{adversarial} {\color{BlueT}StyleGAN} model \citep{stylegan} can also be used for image reconstruction, by applying optimization-based inversion techniques to infer back latents from images. Given an image $x$, they leverage gradient descent to reverse engineer the latent $z$ that gives rise to an output $\hat{x}$ that is as close as possible to the image $x$ while still staying on the model's learned manifold. While these techniques tend to produce samples that share semantic properties with the source images, they oftentimes fail to reconstruct them faithfully. Finally, we compare our model to the \textbf{diffusion-based} {\color{BlueT}DiffAE} \citep{diffae}, which, in contrast to our study, focuses on auto-encoding only, and can be viewed as a predecessor of our approach, as discussed in \cref{related}.

We assess the reconstruction capabilities of the approaches described in this subsection by evaluating a sample set of images produced by their associated public pre-trained checkpointed models.

\subsection{Novel View Synthesis}
For novel view synthesis of 3D objects, we compare SODA to a collection of \textbf{geometry-free and -aware} approaches designed for few-shot settings: {\color{BlueT}PixelNeRF} \citep{pixelnerf} learns to translate a small number of source views into a neural radiance field, and then use volumetric rendering techniques to generate new ones. {\color{BlueT}NeRF-VAE} \citep{nerfvae} extends this idea by leveraging amortized variational inference to learn probablistic neural scene representations. In contrast to these specialized methods, designed specifically for 3D environments, SODA proves considerably more versatile, successfully addressing a broader spectrum of tasks and datasets.

As an alternative to differentiable rendering, geometry-free approaches often use attention mechanisms to directly transform source views into targets: {\color{BlueT}Scene Representation Transformer (SRT)} \citep{srt} parametrizes scenes with the computationally lighter and faster Light-Field formulation \citep{lfn}, and synthesize output views from new perspectives by directly attending to the input views' encodings. The diffusion-based 3DiM \citep{nerd} goes further and makes extensive use of cross-attention throughout all of its network's layers so to directly map sources to targets. In contrast to these approaches, we intentionally introduce a bottleneck into our model that induces a meaningful and compact latent space. This, in turn, offers much tighter control over the model's generative process, opening the door for both semantic manipulation of given scenes, as well as unconditional synthesis of new ones -- two new capabilities that are out of these prior works' reach.

We evaluate the methods described in this subsection either using the authors' official implementations (for NeRF-VAE), or with our own re-implementations (for PixelNeRF and SRT), matching the originally reported performance. 

\subsection{Disentanglement}
In terms of disentanglement, we analyze SODA over a suite of semantically-annotated datasets, and compare it with a series of \textbf{variational} approaches \citep{vae}, which are traditionally known for encouraging the formation of disentangled representations: {\color{BlueT}$\beta$-VAE} \citep{beta-vae}, re-weights the KL regularization term to constrain the latents' capacity; {\color{BlueT}AnnealedVAE} \citep{annealed_vae} slowly relaxes the encoder-decoder bottleneck so to foster gradual learning; {\color{BlueT}FactorVAE} \citep{factor_vae} and {\color{BlueT}$\beta$-TCVAE} \citep{tc_vae} encourage factorization of the latent distribution by reducing the correlations among the axes; {\color{BlueT}DIP-VAE} \citep{dipvae} (variants I and II) penalizes the mismatch between the prior and the posterior, so to similarly encourage factorization within the latter. We evaluate these methods using the official \textit{disentanglement-lib} TensorFlow repository \citep{disen_lib}, while modifying the backbone encoder and decoder architectures to match the ones used in SODA, for better comparability.

\section{Ablation Studies}
\label{ablt}

\begin{table*}[t]
\caption{\textbf{Hyperparameters} of our model, including the encoder, denoiser, linear probe, and view aggregation transformer (for 3D experiments), as well as optimization, sampling and augmentation settings. ($\star$) Depends on the dataset. ($\dagger$) Applied for ImageNet only. ($\ast$) We use a base value of 64 for ImageNet pre-training for downstream classification, and 128 otherwise.} 
\label{tab_impl}
\centering
\scriptsize
\begin{tabular}{lc}
\rowcolor{Blue1} \textbf{Hyperparameter} & \textbf{Value} \\
\rowcolor{Blue2} {\color{DarkBlue} \textbf{Optimization}} & \\
Learning Rate$^\star$ & (1-4)${\times}10^{-4}$ \\
\rowcolor{Blue3} Batch Size$^\star$ & 1024-4096 \\
Learing Rate Schedule$^\dagger$ & Cosine Decay \\
\rowcolor{Blue3} Learing Rate After Decay$^\dagger$ & 0.25$\times$LR \\
Learning Rate Decay Steps$^\dagger$ & $1.2{\times}10^5$ \\
\rowcolor{Blue3} Weight Decay & 0.05 \\
EMA Decay Rate & 0.9999 \\
\rowcolor{Blue3} Warmup Steps & 5000 \\
Gradient Clipping Norm & 0.5 \\
\rowcolor{Blue3} Optimizer & Adam \\
$\beta_1$ & 0.9 \\
\rowcolor{Blue3} $\beta_2$ & 0.95 \\
\rowcolor{Blue2} {\color{DarkBlue} \textbf{Model}} & \\
Latent Dimension$^\star$ & 128-2048 \\
\rowcolor{Blue3} Bottleneck Dropout & 0.1 \\
Classifier-Free Guidance Masking Rate & 0.12 \\
\rowcolor{Blue3} Layer Masking Rate & 0.15 \\
Positional Encoding Dimension & 512 \\
Positional Encoding Scale (\cref{fig_pos_enc}) & 0.0001 \\
\rowcolor{Blue2} {\color{DarkBlue} \textbf{Encoder}} & \\
Architecture & ResNet \\
\rowcolor{Blue3} Size$^\star$ & 18, 50, 50$\times$2 \\
Version & v2 \\
Resolution & 256 (224$^\dagger$) \\
LR multiplier & 2 \\
\rowcolor{Blue3} DropPath \citep{dropath} & 0.1 \\
\rowcolor{Blue2} {\color{DarkBlue} \textbf{View Aggregation Transformer}} & \\
Depth & 2 \\
\end{tabular}
\begin{tabular}{lc}
\rowcolor{Blue1} \textbf{Hyperparameter} & \textbf{Value} \\
\rowcolor{Blue3} Attention Heads Number & 4 \\
Hidden Layer Multiplier & 4 \\
\rowcolor{Blue2} {\color{DarkBlue} \textbf{Denoiser}} & \\
Architecture & UNet \\
Resolution & 128 \\
\rowcolor{Blue3} Base Channels$^\ast$ & 64-128 \\
Channels multipliers & 1,1,2,3,4 \\
\rowcolor{Blue3} Residual blocks per resolution & 2 \\
Selt-Attention resolution & 8,16,32 \\
\rowcolor{Blue3} Attention Head Dimension & 64 \\
Normalization Type & GroupNorm \citep{groupnorm} \\
\rowcolor{Blue3} Dropout Rate & 0.1 \\
\rowcolor{Blue2} {\color{DarkBlue} \textbf{Sampling}} & \\
Classifier-Free Guidance$^\star$ & 2-3 \\
\rowcolor{Blue3} Diffusion Training Steps & 1000 \\
Sampling Strided Steps$^\star$ & 75-250 \\
\rowcolor{Blue2} {\color{DarkBlue} \textbf{Linear Probe}} & \\
Weight Initialization Scale & 0.02 \\
\rowcolor{Blue3} Bias Initialization Scale$^\dagger$ &-10.0 \\
Dropout Rate & 0.1 \\
\rowcolor{Blue3} Augmentation Rate (RandomResizedCrop) & 0.65 \\
\rowcolor{Blue3} Label Smoothing$^\dagger$ & 0.1 \\
\rowcolor{Blue2} {\color{DarkBlue} \textbf{Data Augmentation}} & \\
Augmentation Rate (Cropping+Flipping) & 0.95 \\
\rowcolor{Blue3} Distortation Rate (RandAugment \citep{rand_augment}) & 0.65 \\
Distortion Layers Number & 2 \\
\rowcolor{Blue3} Distortion Magnitude & 9 \\
Distortion Magnitude STD & 0.5 \\
\rowcolor{Blue3} Source View Noise Scale$^\star$ & 0-0.22  \\
\end{tabular}
\end{table*}

To gain better insight into the relative contributions our design decisions make, we conduct thorough ablation and variation studies for each of the model's components, inspecting the (1)~\textbf{feature modulation} used to propagate information between the encoder and the denoiser, (2)~\textbf{data augmentation} strategies for the source and target views, encoding and conditioning schemes of (3)~\textbf{positional information} for our 3D multi-view experiments, (4)~\textbf{sampling configurations} of the denoising process and its classifier-free guidance, and finally, (5)~the encoder and denoiser's respective \textbf{sizes, dimensions and learning rates}.

This study joins ablations presented through the main paper (\cref{cls,disen_quant}) that attest to the strengths and benefits of the model's core aspects and key innovations, like bottleneck compactness (\cref{encoder}), layer modulation (\cref{masking}), redesigned noise schedule (\cref{training}), and incorporation of novel view synthesis as a self-supervised training objective (\cref{views}). 

\subsection{Feature Modulation} 
We explore multiple modulation variants and examine how they fare in terms of generative skills and downstream performance (\cref{tab_ablt_imagenet,tab_celeba_ablt,tab_disen_supp}). As the results suggest, \textbf{modulation-based conditioning} proves considerably more effective than alternative mechanism such as input concatenation $[\bm{x}_t,\bm{x'}]$ (Palette \citep{palette}) or spatial broadcasting (\textit{w/o modulation}) \citep{spatial_broadcaster}, with respective deltas of 15.5\% and 12.3\% at classification over ImageNet (top1) and CelebA (F1), and 0.32 (out of 1.0) mean SSIM improvement at novel view synthesis. \textbf{Layer modulation} proves beneficial too, enhancing disentanglement scores, with up to 13.9\% improvement, and generative capabilities, with 0.13 increase in SSIM and halving of LPIPS for ImageNet reconstructions.

We further assess ways to integrate the guidance of the timestep $t$ and the latent $\bm{z}$, and as an alternative to our \textbf{two-stage guidance} approach, where the denoiser's activations $\bm{h}$ are modulated first by $t$ and subsequently by $\bm{z}$, we map them instead to a single pair $(\bm{w}_s, \bm{w}_b)$ either through summation or concatenation (\textit{sum/concat mod.}) which is then used to modulate the activations:
$\text{AdaGN}(\bm{h},t,\bm{z})={\bm{w}_s}\text{\text{GroupNorm}}(\bm{h})+{\bm{w}_b}$ 
However, our two-step strategy proves stronger than this variant. Likewise, \textbf{scaling the features multiplicatively} with $\bm{z}_s$, as opposed to adding a bias term $\bm{z_b}$ only, leads to small improvements across different datasets.

\subsection{Data Augmentation}
We examine the impact of data augmentations on the model's performance, and analyze variations of the augmentation method itself as well as the inputs it is applied to (\cref{fig_plot}). At training, our model receives two inputs: a clean view $\bm{x'}$ processed by the encoder $\mathcal{E}$, and a noisy view $\bm{x}_t$ denoised by the decoder $\mathcal{D}({\bm{x}_t}{\mid}\bm{x'})$, aiming to recover $\bm{x}=\bm{x}_0$. With the exception of native multi-view datasets (e.g. ShapeNet), we create the source and target views $\bm{x'}$ and $\bm{x}_t$ by applying random data augmentations at each training step on the original image $\bm{x}$ (from the dataset). 

We test the impact of applying augmentations either just to the source view, just to the target view, to both, or to none of them. We observe that \textbf{augmenting the source} is more critical than the target in terms of its influence on downstream classification performance, and that the model still achieves 55.1 when the source and the target views remain equal. Moreover, we find it valuable to add \textbf{low Gaussian noise} to the source views read by the encoder, yielding 1.3\% improvement in ImageNet classification accuracy and improving LPIPS scores relatively by 33\%. 

\subsection{Pose Conditioning}

\begin{table}[t]
\caption{\textbf{Ablations on ShapeNet}, varying the \textbf{classifier-free guidance} settings: either masking the latent $\bm{z}$ that encodes the source image view, masking the pose information $\bm{r}$ of the source and target views, or independently masking both.} 
\label{tab_pos_masking}
\centering
\scriptsize
\begin{tabular}{lcccc}
\rowcolor{Blue1} \textbf{Masking} & \textbf{PSNR $\uparrow$} & \textbf{SSIM $\uparrow$} & \textbf{FID $\downarrow$} & \textbf{LPIPS $\downarrow$} \\
\rowcolor{Blue2} \multicolumn{5}{l}{\color{DarkBlue} \textbf{ShapeNet}} \\
Latent & \textbf{27.42} & \textbf{0.947} & \textbf{0.74} & \textbf{0.039} \\
\rowcolor{Blue3} Pose & 27.16 & 0.940 & 0.96 & 0.041 \\
Latnet + Pose & 27.11 & 0.938 & 0.95 & 0.041 \\
\rowcolor{Blue2} \multicolumn{5}{l}{\color{DarkBlue} \textbf{GSO}} \\
Latent & 24.12 & 0.937 & 2.22 & 0.065 \\
\rowcolor{Blue3} Pose & 24.25 & 0.939 & 2.48 & 0.062 \\
\rowcolor{Blue1} Latnet + Pose & \textbf{24.97} & \textbf{0.945} & \textbf{1.51} & \textbf{0.054}
\end{tabular}
\end{table}

We compare different encoding schemes of the camera perspectives for the 3D novel view synthesis task (\cref{tab_pos_ablt}). Given a camera pose $\bm{p}$, we can use a closed-form calculation to derive a 2D grid of rays $\bm{r}=(\bm{o},\bm{d})$ of dimension $H{\times}W{\times}6$ with origins $\bm{o}$ and directions $\bm{d}$. We can then represent each ray through concatenation: $[\bm{o},\bm{d}]$ (\textit{concat}), as commonly done in prior works \citep{pixelnerf,srt}, or instead, express them with a parametric sum: $\bm{o}\,+\,s_d{\cdot}\bm{d}$, where $s_d$ is a scaling factor that can be chosen in different ways: either normalizing $\bm{d}$ to a length of 1, casting it onto the image plane, or, as we propose, on a sphere that centers at the object, i.e. the origin (\textit{Normalized}, \textit{Plane} and \textit{Sphere}). We can further describe the rays either using Polar or Cartesian coordinates, embedded with sinusoidal positional encoding \citep{transformer} as explained above (\cref{supp_impl}). We compare these alternatives, and find that \textbf{casting the rays on a sphere} performs most effectively, and that for this case, \textbf{Polar coordinates} outperform the Cartesian ones.

We further experiment with representing the camera pose as a single vector $\bm{p}$ that captures its position and direction in Polar coordinates, either considering $\bm{p}_t-\bm{p}_s$, the relative camera transformation from the source to the target, or concatenating the two absolute viewpoints $[\bm{p}_s,\bm{p}_t]$. We then encode the information with sinusoidal positional encoding \citep{transformer}), and use the resulting vector to guide the denoiser's operation through feature modulation, similarly to the latent $\bm{z}$. However, the ablations show that integrating the camera perspective by concatenating a 2D grid of rays surpasses both the modulation-based pose conditioning as well as a hybrid alternative that simultaneously uses both techniques.

Finally, we experiment with different masking techniques as part of the classifier-free guidance (\cref{views}), either randomly masking the latent representation $\bm{z}$ that encodes the source image view, masking the pose information (namely, the rays 2D grid $\bm{r}$), or independently masking both. We interestingly note that the ideal masking vary for different datasets: while masking of both the latent and pose improves performance for the real-world Google Scanned Objects, it reduces the performance for ShapeNet. Qualitatively, masking both the pose $\bm{r}$ and the latent $\bm{z}$ enhances the model's generative flexibility, allowing it to synthesize either novel objects at requested camera perspectives, or arbitrary novel views even at the absence of source or target's pose information (supplementary figures will be added very soon).

\subsection{Sampling Configuration}
We vary the guidance strength $g$ and timesteps striding $l$ (i.e. number of timesteps used at sampling) and analyze their impact on the generated images' quality along different metrics (\cref{fig_guidance_sampling}). The model is robust to variation in both settings, with optimal values commonly achieved at $g=2$ and $l=150$ (considering different metrics and datasets). Classifier-free guidance consistently yields higher-quality images than unguided sampling, while too strong guidance (like ${\geq}5$) results in a slight reduction in scores and potential visual artifacts.

As per the number of sampling steps, while PSNR and LPIPS scores tend to remain constant, we interestingly observe an inverse correlation between the process length's influence on FID vs. SSIM, the former reflecting sharpness and fidelity while the latter capturing similarity to the target: Sampling images over more steps tends to improve their realism, but may simultaneously induce subtle variations, as the samples begin to slightly move away from the mean estimated target. As aforementioned, we find that $l=150$ offers a favorable balance between these two qualities.

\end{document}